\documentclass[lettersize,journal]{IEEEtran}
\pdfoutput=1
\usepackage{amsmath,amsfonts}
\usepackage{algorithmic}
\usepackage{algorithm}
\usepackage{array}
\usepackage[caption=false,font=normalsize,labelfont=sf,textfont=sf]{subfig}
\usepackage{textcomp}
\usepackage{stfloats}
\usepackage{url}
\usepackage{verbatim}
\usepackage{graphicx}
\usepackage{cite}
\usepackage{hyperref}       
\usepackage{booktabs}       
\usepackage{nicefrac}       
\usepackage{microtype}      
\usepackage{xcolor}         
\usepackage{multirow}
\usepackage{caption}
\usepackage{epsfig}
\usepackage{amsmath}
\usepackage{amssymb}
\usepackage{amsthm}
\usepackage{subfloat}
\usepackage{times}
\usepackage{threeparttable}

\newtheorem{remark}{Remark}
\newtheorem{definition}{Definition}

\usepackage{xcolor}
\definecolor{orange}{rgb}{255,107,0}

\begin{document}

\title{DTR: A Unified Deep Tensor Representation Framework for Multimedia Data Recovery}

\author{Ting-Wei Zhou, Xi-Le Zhao*, Jian-Li Wang, Yi-Si Luo, Min Wang, Xiao-Xuan Bai, \\ and Hong Yan \IEEEmembership{}
\thanks{Ting-Wei Zhou and Xi-Le Zhao are with the School of Mathematical
	Sciences/Research Center for Image and Vision Computing, University of
	Electronic Science and Technology of China, Chengdu, Sichuan 611731,
	China (e-mail: zhoutingwei2021@163.com; xlzhao122003@163.com).}
\thanks{Jian-Li Wang is with the School of Computing and Artificial Intelligence, Southwest Jiaotong University, Chengdu, Sichuan 611756, China (e-mail: wangjianli\_123@163.com).}
\thanks{Yi-Si Luo is with the School of Mathematics and Statistics, Xi’an Jiaotong	University, Xi’an 710049, China (e-mail: yisiluo1221@foxmail.com).}
\thanks{Min Wang is with the School of Information Management, Jiangxi University of Finance and Economics, Nanchang, Jiangxi 330013, China (e-mail: minwang1989@126.com).}
\thanks{Xiao-Xuan Bai is with the Glasgow College, University of
	Electronic Science and Technology of China, Chengdu, Sichuan 611731,
	China (e-mail: abnxnxa@outlook.com).}
\thanks{Hong Yan is with the Department of Electrical Engineering, City University of Hong Kong, Kowloon, Hong Kong SAR (e-mail: h.yan@cityu.edu.hk).}
}

\markboth{}%
{Shell \MakeLowercase{\textit{et al.}}: A Sample Article Using IEEEtran.cls for IEEE Journals}

\IEEEpubid{}

\maketitle

\begin{abstract}
	Recently, the transform-based tensor representation has attracted increasing attention in multimedia data (e.g., images and videos) recovery problems, which consists of two indispensable components, i.e., transform and characterization. Previously, the development of transform-based tensor representation mainly focuses on the transform aspect. Although several attempts consider using shallow matrix factorization (e.g., singular value decomposition
	and negative matrix factorization) to characterize the frontal
	slices of transformed tensor (termed as latent tensor), the faithful characterization aspect is underexplored. To address this issue, we propose a unified \textbf{D}eep \textbf{T}ensor \textbf{R}epresentation (termed as DTR) framework by synergistically combining the deep latent generative module and the deep transform module. 
 	Especially, the deep latent generative module can faithfully generate the latent tensor as compared with shallow matrix factorization. The new DTR framework not only allows us to better understand the classic shallow representations, but also leads us to explore new representation. 
	 To examine the representation ability of the proposed DTR, we consider the representative multi-dimensional data recovery task and suggest an unsupervised DTR-based multi-dimensional data recovery model. Extensive experiments demonstrate that DTR achieves superior performance compared to state-of-the-art methods in both quantitative and qualitative aspects, especially for fine details recovery.
\end{abstract}


\section{Introduction}
\IEEEPARstart{M}{ultimedia} data, such as color images \cite{huang2021quaternion, liu2018image, fang2019variational}, multispectral images (MSIs) \cite{anandkumar2014tensor, gandy2011tensor}, magnetic resonance imaging (MRI) images \cite{fabian2021data}, and videos \cite{li2017low,tian2018snowflake}, is commonly encountered in various real-world applications, including but not limited to agricultural monitoring \cite{you2017deep}, hyperspectral image (HSI) restoration \cite{li2020superpixel, zeng2022multimodal}, and medical imaging \cite{lv2021discriminant}. However, the acquired multimedia data is usually incomplete owing to sensor failures or other abnormal conditions that occur during the acquisition and transmission process, which can severely hinder subsequent applications \cite{chen2020neural, chen2021deep, chen2021efficient}. The task of estimating missing data from the observed data can be formulated as the multimedia data recovery problem. Besides, with the development of multimedia technology, multimedia data has become increasingly multi-dimensional. Hence, it is important to develop effective multi-dimensional data recovery methods.

	\begin{figure}[t]
	\centering
	\subfloat{
		\begin{minipage}[b]{0.194\linewidth}
			\captionsetup{
				font={small},     
				labelfont=bf,        
				textfont={rm}, 
				singlelinecheck=true 
			}
			\caption*{Observed}

			\includegraphics[width=1\linewidth]{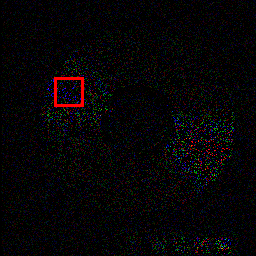}
			\put(-20.1,0){\includegraphics[scale=0.8]{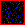}}
			\vspace{-0.2cm}
			\caption*{{\fontsize{6}{7} \selectfont PSNR 18.42 dB}}
			\includegraphics[width=1\linewidth]{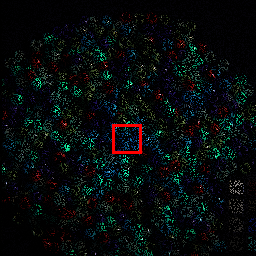}
			\put(-20.1,0){\includegraphics[scale=0.8]{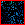}}
			\vspace{-0.2cm}
			\caption*{{\fontsize{6}{7} \selectfont PSNR 15.77 dB}}
	\end{minipage}}
	\hspace{-2mm}
	\subfloat{
		\begin{minipage}[b]{0.194\linewidth}
			\captionsetup{
				font={small},     
				labelfont=bf,        
				textfont={rm}, 
				singlelinecheck=true 
			}
			\caption*{t-SVD}

			\includegraphics[width=1\linewidth]{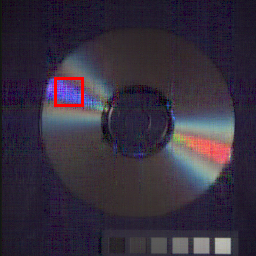}
			\put(-20.1,0){\includegraphics[scale=0.8]{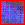}}
			\vspace{-0.2cm}
			\caption*{{\fontsize{6}{7} \selectfont PSNR 30.01 dB}}
			\includegraphics[width=1\linewidth]{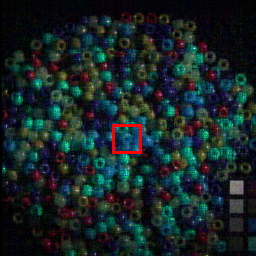}
			\put(-20.1,0){\includegraphics[scale=0.8]{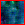}}
			\vspace{-0.2cm}
			\caption*{{\fontsize{6}{7} \selectfont PSNR 23.41 dB}}
	\end{minipage}}
	\hspace{-2mm}
	\subfloat{
		\begin{minipage}[b]{0.194\linewidth}
				\captionsetup{
				font={small},     
				labelfont=bf,        
				textfont={rm}, 
				singlelinecheck=true 
			}
			\caption*{HLRTF}

			\includegraphics[width=1\linewidth]{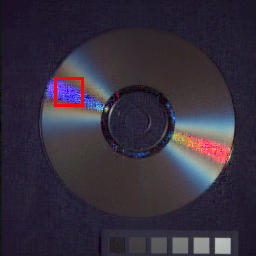}
			\put(-20.1,0){\includegraphics[scale=0.8]{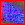}}
			\vspace{-0.2cm}
			\caption*{{\fontsize{6}{7} \selectfont PSNR 31.91 dB}}
			\includegraphics[width=1\linewidth]{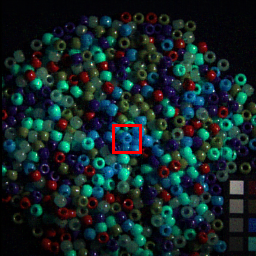}
			\put(-20.1,0){\includegraphics[scale=0.8]{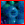}}
			\vspace{-0.2cm}
			\caption*{{\fontsize{6}{7} \selectfont PSNR 30.41 dB}}
	\end{minipage}}
	\hspace{-2mm}
	\subfloat{
		\begin{minipage}[b]{0.194\linewidth}
			\captionsetup{
				font={small},     
				labelfont=bf,        
				textfont={rm}, 
				singlelinecheck=true 
			}
			\caption*{DTR}

			\includegraphics[width=1\linewidth]{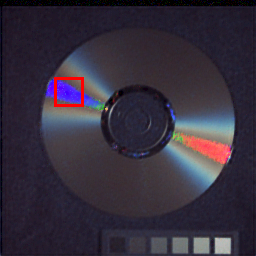}
			\put(-20.1,0){\includegraphics[scale=0.8]{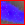}}
			\vspace{-0.2cm}
			\caption*{{\fontsize{6}{7} \selectfont \textbf{PSNR 33.95 dB}}}
			\includegraphics[width=1\linewidth]{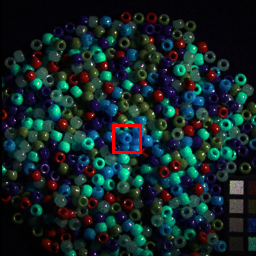}
			\put(-20.1,0){\includegraphics[scale=0.8]{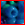}}
			\vspace{-0.2cm}
			\caption*{{\fontsize{6}{7} \selectfont \textbf{PSNR 33.38 dB}}}
	\end{minipage}}
	\hspace{-2mm}
	\subfloat{
		\begin{minipage}[b]{0.194\linewidth}
			\captionsetup{
				font={small},     
				labelfont=bf,        
				textfont={rm}, 
				singlelinecheck=true 
			}
			\caption*{Original}

			\includegraphics[width=1\linewidth]{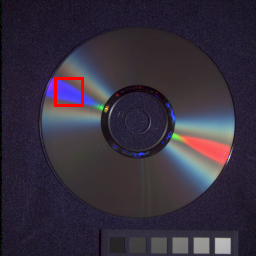}
			\put(-20.1,0){\includegraphics[scale=0.8]{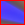}}
			\vspace{-0.2cm}
			\caption*{{\fontsize{6}{7} \selectfont PSNR Inf}}
			\includegraphics[width=1\linewidth]{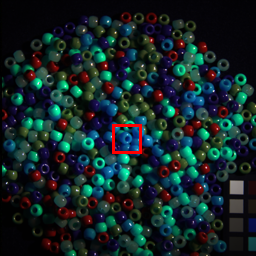}
			\put(-20.1,0){\includegraphics[scale=0.8]{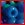}}
			\vspace{-0.2cm}
			\caption*{{\fontsize{6}{7} \selectfont PSNR Inf}}
	\end{minipage}}
	\vspace{-0.2cm}
	\caption{The recovered pseudo-color images (R:10, G:20, B:30) by shallow characterization-based methods, i.e., t-SVD (based on SVD)  \cite{zhang2014novel} and HLRTF (based on negative matrix factorzation) \cite{luo2022hlrtf}, and deep characterization-based method, i.e., the proposed DTR.} 
	
	\label{visual}
\end{figure}

Recently, the transform-based tensor representation has emerged as a powerful tool for multi-dimensional data recovery problems, as demonstrated by extensive studies \cite{zhao2020deep, jiang2020multi, luo2022hlrtf,jiang2020framelet, lu2019low, madathil2018dct}. The transform-based tensor representation represents the multi-dimensional data as an implicit low-rank tensor and exploits its low-rankness in the transform domain. In the transform-based tensor representation, two indispensable components are (i) the transform that captures the frontal slice relationships of multi-dimensional data \cite{jiang2020framelet, lu2019low, madathil2018dct}, and (ii) the characterization of the frontal slices of transformed tensor (termed as latent tensor) \cite{kilmer2011factorization, zhou2017tensor, wu2021fast}. 

Previously, the development of transform-based tensor representation mainly focuses on the transform perspective, such as the discrete Fourier transform (DFT) \cite{kilmer2011factorization,lu2019tensor}, the discrete cosine transform (DCT) \cite{madathil2018dct}, unitary transform \cite{ng2020patched, song2020robust}, non-invertible transform \cite{jiang2020framelet, jiang2021dictionary, kong2021tensor}, and nonlinear transform \cite{luo2022hlrtf}. However, few methods have paid attention to the second component, i.e., the characterization of the frontal slices of latent tensor. To date, the transform-based tensor representation usually adopts shallow matrix factorization to characterize the frontal slices of latent tensor, such as singular value decomposition (SVD) \cite{kilmer2011factorization}, negative matrix factorization (NMF) \cite{luo2022hlrtf,zhou2017tensor}, and QR decomposition \cite{zheng2021tensor, wu2021fast}. Due to the limited characterization ability, these shallow methods may not be expressive enough to faithfully characterize the latent tensor. Moreover, 
shallow matrix factorization characterizes the intra-slice relationships of latent tensor separately, which neglects the inter-slice relationships of latent tensor. 
\IEEEpubidadjcol

To address the above issues, we propose a unified \textbf{D}eep \textbf{T}ensor \textbf{R}epresentation (termed as DTR) framework by synergistically combining the deep latent generative module and the deep transform module. Specifically, we suggest a deep latent generative module to faithfully generate the latent tensor by characterizing both intra-slice and inter-slice relationships of latent tensor. Meanwhile, we utilize an untrained neural network applied on the third dimension of latent tensor as the deep transform module to capture the frontal slice relationships of multi-dimensional data. The new DTR framework not only allows us to better understand the classic shallow representations, but also leads us to explore new representation. To examine the representation ability of the proposed DTR, we consider the representative multi-dimensional data recovery task and suggest an unsupervised DTR-based multi-dimensional data recovery model. Thanks to the superior characterization ability of deep latent generator, our method can obtain better multi-dimensional data recovery results than shallow characterization-based methods; as shown in Fig. \ref{visual}.


The main contributions of this study are: 
\begin{itemize}
	\item To overcome the limitations of shallow characterization in the transform-based tensor representation, we propose a unified deep tensor representation framework, i.e., DTR, by synergistically combining the deep latent generative module and the deep transform module. The new DTR framework not only allows us to better understand the classic shallow representations, but also leads us to explore new representation.
	
	\item Based on the DTR, we propose an unsupervised multi-dimensional data recovery model. Experimental results demonstrate that the proposed method can achieve significant performance improvements over state-of-the-art methods, especially for fine details recovery. 
\end{itemize} 

This paper is organized as follows. Section \ref{section:2} introduces necessary notations and preliminaries. Section \ref{section:3} provides a review of related work. Section \ref{section:4} presents the proposed DTR. Section \ref{section:5} and \ref{section:6} present the results of extensive experiments and discussions. Finally, we conclude this work in Section \ref{section:7}.

\section{Notations and preliminaries}
\label{section:2}

\subsection{Notations}
Matrices and tensors are denoted by capitalized letters, e.g., $\mathbf{X}$ and capitalized calligraphic letters, e.g., $\mathcal{X}$, respectively. For a matrix  $\mathbf{X} \in \mathbb{R}^{n_{1} \times n_{2}}$, we use $\mathbf{X}(i,j)$ to denote the $(i, j)$-th element of $\mathbf{X}$. Meanwhile, for a third-order tensor $\mathcal{X} \in \mathbb{R}^{n_{1} \times n_{2} \times n_{3}}$, we use $\mathcal{X}(i,j,k)$ to denote the $(i, j, k)$-th element of $\mathcal{X}$. $\mathcal{X}^{(k)} \in \mathbb{R}^{n_{1} \times n_{2}}$ denotes the $k$-th frontal slice of $\mathcal{X}$. $\odot$ denotes the point-wise multiplication (i.e., Hadamard product). The Frobenius norm of tensor $\mathcal{X}$ is defined as $\|\mathcal{X}\|_F=\sqrt{\sum_{i,j,k}\left|\mathcal{X}(i,j,k)\right|^2}$.  $\Delta$ denotes the face-wise product between two tensors \cite{kilmer2013third}. Given $\mathcal{X}\in \mathbb{R}^{n_{1} \times n_{2} \times n_{3}}$, $\mathcal{Y}\in \mathbb{R}^{n_{2} \times n_{4} \times n_{3}}$, the face-wise product between $\mathcal{X}$ and $\mathcal{Y}$ is $\mathcal{X}\Delta\mathcal{Y} = \mathcal{Z} \in \mathbb{R}^{n_{1} \times n_{4} \times n_{3}}$, where $\mathcal{Z}^{(k)}=\mathcal{X}^{(k)}\mathcal{Y}^{(k)} (k = 1, \ldots, n_3)$.

\subsection{Preliminaries} 

\begin{definition}[Mode-3 Unfolding \cite{doi:10.1137/07070111X}] For a third-order tensor $\mathcal{A}\in \mathbb{R}^{n_1 \times n_2 \times n_3}$, its mode-$3$ unfolding $\mathcal{A}_{(3)}$ is a $n_3 \times n_1 n_2$ matrix, which satisfies that $\mathcal{A}_{(3)}(i_3, j)=\mathcal{A}(i_1, i_2, i_3)$, where $j=1+\sum_{s=1}^{2}\left(i_{s}-1\right)\textit{J}_{s}$ with $\textit{J}_{s}=\prod_{m=1}^{s-1}n_m$.
	\label{unfold}
\end{definition}

\begin{definition}[Mode-3 Tensor-Matrix Product \cite{doi:10.1137/07070111X}] The mode-3 tensor-matrix product of a $n_1 \times n_2 \times n_3$ tensor $\mathcal{X}$ and a $J \times n_3$ matrix $\mathbf{A}$ is a $n_1 \times n_{2} \times J $ tensor denoted by $\mathcal{X} \times_3 \mathbf{A}$ and satisfied 
	\begin{equation}
		\left(\mathcal{X} \times_3 \mathbf{A}\right)(i_1, i_2, j)=\sum_{i_3=1}^{n_3} \mathcal{X}(i_1, i_2, i_3) \mathbf{A}(j, i_3).
	\end{equation}
\end{definition}

According to above mentioned two definitions, we have
\begin{equation}
	\mathcal{Y}=\mathcal{X} \times_3 \mathbf{A} \Leftrightarrow \mathcal{Y}_{(3)}=\mathbf{A} \mathcal{X}_{(3)}.
\end{equation}

\begin{definition}[Tensor Tubal-Rank \cite{kilmer2013third}] The tensor tubal-rank of $\mathcal{A} \in \mathbb{R}^{n_{1} \times n_{2} \times n_{3}}$ is defined as
	\begin{equation}
		rank_t(\mathcal{A})\triangleq \max \limits_{k=1,2,\ldots,n_{3}}{rank\left((\mathcal{A}\times_3 \mathbf{F})^{(k)}\right)},
	\end{equation}
	where $\mathbf{F} \in \mathbb{R}^{n_{3} \times n_{3}}$ is the DFT matrix.
\end{definition}

\begin{definition}
			[T-Product \cite{kilmer2013third}] The tensor-tensor product
			between $\mathcal{A} \in \mathbb{R}^{n_{1} \times n_{2} \times n_{3}}$, and $\mathcal{B} \in \mathbb{R}^{n_{2} \times n_{4} \times n_{3}}$ is defined as
			$\mathcal{A} * \mathcal{B}=((\mathcal{A}\times_3 \mathbf{F})\Delta(\mathcal{B}\times_3\mathbf{F}))\times_3 \mathbf{F}^{-1} \in \mathbb{R}^{n_{1} \times n_{4} \times n_{3}}$, where
			$\mathbf{F} \in \mathbb{R}^{n_{3} \times n_{3}}$ is the DFT matrix, and $\mathbf{F}^{-1}$ is the inverse DFT matrix.
			\label{d}
\end{definition}

	\section{Related work}
	
		In this section, we briefly review the transform-based tensor representation, with a focus on the advances of two indispensable components, i.e., transform and characterization.
	\label{section:3}
	\subsection{The Transform Perspective}
	
	An indispensable component in the transform-based tensor representation is the transform that captures the frontal slice relationships of multi-dimensional data. Over the past decade, researchers have explored various transforms \cite{madathil2018dct, song2020robust, jiang2020framelet, kong2021tensor, luo2022hlrtf} to facilitate the most classic transform-based tensor representation, i.e., tensor singular value decomposition (t-SVD). For instance, Kilmer \textit{et al.} \cite{kilmer2011factorization} first introduced the DFT along the third dimension under the t-SVD framework. Madathil \textit{et al.} \cite{madathil2018dct} adopted the real-valued DCT to avoid the high computational cost caused by complex operations in DFT-based original t-SVD. 
	Lu \textit{et al.} \cite{lu2019low} deduced the new tensor tubal rank based on the invertible linear transform and provided the theoretical bound for the exact recovery under certain tensor incoherence conditions for tensor completion. Unlike previous invertible transforms, Jiang \textit{et al.} \cite{jiang2020framelet} employed the non-invertible framelet transform to construct a framelet latent tensor. In addition to the previously pre-defined transforms, some data-driven transforms have emerged promising results in multi-dimensional data recovery problems. For example, Song \textit{et al.} \cite{song2020robust} constructed a data-dependent unitary transform, which can be obtained by taking the SVD of the unfolding matrix. Kong \textit{et al.} \cite{kong2021tensor} defined the new tensor Q-rank based on novel data-dependent transform, which can be learned by principal component analysis. Luo \textit{et al.} \cite{luo2022self} proposed a nonlinear transform-based tensor nuclear norm (TNN), where the nonlinear transform is obtained by learning a multi-layer neural network solely using the observed tensor in a self-supervised manner.
	\subsection{The Characterization Perspective}
	Another indispensable component in the transform-based tensor representation is the characterization of the frontal slices of latent tensor. The original t-SVD \cite{kilmer2011factorization} adopted the standard matrix SVD to characterize the frontal slices of latent tensor. Zhou \textit{et al.} \cite{zhou2017tensor} introduced NMF as an alternative approach to efficiently characterize the low-rankness of the frontal slices of latent tensor. In addition, to further enhance the accuracy and efficiency of the characterization, later work \cite{zheng2021tensor, wu2021fast} adopted an approximate SVD method based on QR decomposition to characterize the frontal slices of latent tensor, which not only reduces the computational complexity but also yields superior results compared to SVD-based method \cite{zhang2014novel}. Despite the benefits of shallow matrix factorization (such as SVD, NMF, and QR decomposition), they may not be representative enough to characterize the latent tensor. Moreover, shallow matrix factorization characterizes the intra-slice relationships of latent tensor separately, which neglects the inter-slice relationships of latent tensor.
	
	\section{Proposed method}
	\label{section:4}
		In this section, we first introduce the proposed unified deep tensor representation framework, i.e., DTR. Then, we discuss the relationships between the proposed DTR and classic shallow representations. 
	Finally, we give the unsupervised DTR-based multi-dimensional data recovery model.
	
	\subsection{The Proposed DTR}
\begin{figure*}[t]
	\centering
	\includegraphics[width=\hsize]{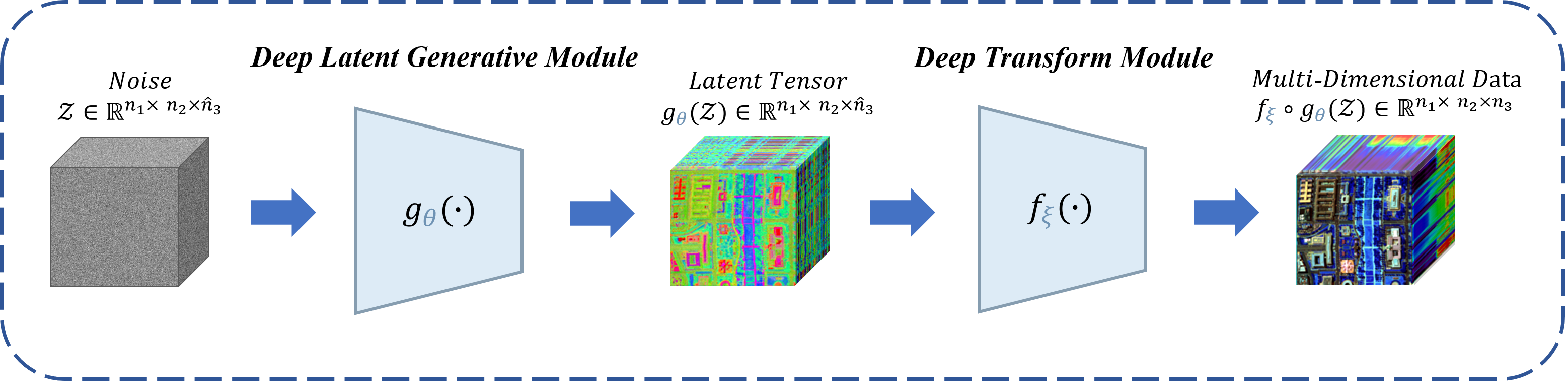}
	\caption{Diagram of our DTR framework\protect\footnotemark[1]. The deep latent generative module $g_\theta(\cdot)$ generates the latent tensor and the deep transform module $f_\xi(\cdot)$ captures the frontal slice relationships of multi-dimensional data.
	}
	\label{fig1}
\end{figure*}

	Inspired by the transform-based tensor representation, we propose a unified deep tensor representation framework, i.e., DTR, by synergistically combining the deep latent generative module and the deep transform module. The diagram of our DTR framework is shown in Fig. \ref{fig1}, and the mathematical formulation can be expressed as follows:
	\begin{equation}
		\mathcal{X}=f_\xi \circ g_\theta(\mathcal{Z}),
		\label{e5}
	\end{equation}  
	where $g_\theta(\cdot)$ is the deep latent generative module, $\theta$ denotes the network parameter of $g(\cdot)$, and $\mathcal{Z}$ is a random noise. $f_\xi(\cdot)$ is the deep transform module, $\xi$ denotes the network parameter of $f(\cdot)$, $\circ$ denotes the composition of functions, and $\mathcal{X}\in \mathbb{R}^{n_{1} \times n_{2} \times n_{3}}$ is the multi-dimensional data represented by the proposed DTR. Two indispensable components in the proposed DTR framework, i.e., the deep latent generative module $g_\theta(\cdot)$ and the deep transform module $f_\xi(\cdot)$, complement to each other and work together to make up a unified framework. Now, we detailedly introduce the two indispensable components in our DTR framework. 
	\subsubsection{Deep Latent Generative Module}
	Previous representation usually adopts shallow matrix factorization to characterize the intra-slice relationships of latent tensor separately, which neglects the inter-slice relationships of latent tensor. In our DTR framework, we suggest the deep latent generative module $g_\theta(\cdot)$
	to generate the latent tensor, which can be formulated as follows:  
	\begin{equation}
		\mathcal{\hat{X}}=g_\theta(\mathcal{Z}),
		\label{e2}
	\end{equation}
	where $g_\theta(\cdot)$: $\mathbb{R}^{n_{1} \times n_{2} \times \hat{n}_{3}}\rightarrow \mathbb{R}^{n_{1} \times n_{2} \times \hat{n}_{3}}$, $\theta$ denotes the network parameter of $g(\cdot)$, $\mathcal{Z}\in \mathbb{R}^{n_{1} \times n_{2} \times \hat{n}_{3}}$ is a random noise, and $\mathcal{\hat{X}}\in \mathbb{R}^{n_{1} \times n_{2} \times \hat{n}_{3}}$ represents the latent tensor.
	Here, $g_\theta(\cdot)$ refers to an untrained neural network that maps a random noise into the latent tensor.
	
	Compared with shallow matrix factorization (e.g., SVD and NMF), the deep latent generator can faithfully generate the latent tensor due to the strong representation ability of untrained neural network. In addition, the deep latent generator can faithfully characterize both intra-slice and inter-slice relationships of latent tensor.
	Based on these two advantages, the deep latent generator has superior characterization ability than shallow matrix factorization, rationally leading to better performance for recovering the fine details of multi-dimensional data (see examples in Fig. \ref{visual}). 
	
	\subsubsection{Deep Transform Module}Now, we are at the position to generate the multi-dimensional data from the latent tensor. To better capture the frontal slice relationships of multi-dimensional data, we utilize an untrained neural network $f_\xi(\cdot)$
	applied on the third dimension of latent tensor as the deep transform module, i.e., 
	\begin{equation}
		\mathcal{X}=f_\xi(\mathcal{\hat{X}}),
		\label{e4}
	\end{equation}
where $f_\xi(\cdot)$: $\mathbb{R}^{n_{1} \times n_{2} \times \hat{n}_{3}}\rightarrow \mathbb{R}^{n_{1} \times n_{2} \times n_{3}}$, $\xi$ denotes the network parameter of $f(\cdot)$, $\mathcal{\hat{X}}\in \mathbb{R}^{n_{1} \times n_{2} \times \hat{n}_{3}}$ is the output of $g_\theta(\mathcal{Z})$, i.e., the latent tensor, and $\mathcal{X}\in \mathbb{R}^{n_{1} \times n_{2} \times n_{3}}$ is the multi-dimensional data represented by the proposed DTR. Here, $f_\xi(\cdot)$ maps the latent tensor into the original domain to capture the frontal slice relationships of multi-dimensional data. 

	
The proposed DTR framework allows us to more deeply understand existing representations. Now, we discuss the relationships between the proposed DTR and classic shallow representations. 

\begin{remark}
Let $\{\mathcal{W}_m \in \mathbb{R}^{r_{m} \times r_{m-1} \times n_{3} }\}_{m=1}^{L}$ (with $r_{L}=n_{1}$, $r_{0}=n_{2}$). Suppose that $g_\theta(\cdot)$ is given as follows:
\begin{equation}
		g_\theta(\mathcal{Z})=\mathcal{W}_{L}\Delta\sigma\left(\mathcal{W}_{L-1}\Delta\cdots\mathcal{W}_3\Delta\sigma\left(\mathcal{W}_2\Delta\mathcal{W}_{1}\Delta\mathcal{Z}\right)\right),		
	\label{e6}
\end{equation}
where each frontal slice of $\mathcal{Z}$ is set as an identity matrix, $\Delta$ denotes the face-wise product between two tensors \cite{kilmer2013third}, and $\sigma(\cdot)$ is a nonlinear activation function.   
	 The proposed DTR can degenerate into classic shallow representations:
	 
	 (i) When $L>2$ and $f_\xi(\cdot)$ is set as a identity mapping, our DTR degenerates into the deep matrix factorization \cite{li2022adaptive, fan2018matrix}, which is applied on each frontal slice of the tensor separately. 
	
	(ii) When $L=2$ and $f_\xi(\cdot)$ is set as a deep fully connected network (FCN), our DTR degenerates into the hierarchical low-rank tensor factorization \cite{luo2022hlrtf}.
	
	
	(iii) When $L=2$ and $f_\xi(\cdot)$ is set as an inverse DFT, our DTR degenerates into the classic low-tubal-rank tensor factorization \cite{zhou2017tensor}.
	
	\label{r1}	
\end{remark}
\footnotetext[1]{If we start from the original tensor, we need to calculate the latent tensor via one transform and then go back to the original tensor via the inverse transform under the invertibility assumption. Here, we start from the latent tensor, which only needs to go back to the original tensor via one transform without the invertibility assumption.} 
Moreover, the proposed DTR framework also allows us to explore new representation. Previously, the faithful characterization aspect is underexplored. Several attempts consider using shallow matrix factorization (e.g., SVD and NMF) in the transform-based tensor representation to characterize the intra-slice relationships of latent tensor separately, which neglects the inter-slice relationships of latent tensor. To faithfully characterize both intra-slice and inter-slice relationships of latent tensor, we can generate the latent tensor with an untrained neural network by leveraging its expressive power under the DTR framework; see Fig. \ref{visual} for the comparison between shallow characterizations and deep characterization in the DTR. 

	\subsection{Unsupervised Multi-Dimensional Data Recovery Model}
	To examine the representation ability of the proposed DTR, we consider the representative multi-dimensional data recovery task and suggest an unsupervised DTR-based multi-dimensional data recovery model, as follow:
	\begin{equation}
		\min\limits_{\xi, \theta} \|\mathcal{M}\odot(\mathcal{X}-\mathcal{O})\|_{F}^2,\\ 
		\  \text{where} \quad \mathcal{X}=f_\xi \circ g_\theta(\mathcal{Z}). 
		\label{e9}
	\end{equation}
	Here, $\mathcal{O}$ is the observed multi-dimensional data, $\mathcal{X}$ is the multi-dimensional data represented by the proposed DTR, $\mathcal{M}$ is a binary mask tensor that sets the observed location to be ones and others be zeros, $g(\cdot)$ is the deep latent generative module, and $f_\xi(\cdot)$ is the deep transform module. 

	
	Although the model (\ref{e9}) is highly non-convex and nonlinear, there are still numerous available methods that can solve it. Specifically, we consider using the popular gradient descent optimization algorithm in deep learning, i.e., the adaptive moment estimation (Adam) \cite{kingma2014adam} algorithm, to
	effectively optimize $\xi$ and $\theta$. Once the optimal $\xi$ and $\theta$ are obtained, the multi-dimensional data $\mathcal{X}$ can be achieved by Eq. (\ref{e5}).

		\begin{table*}[t]\scriptsize
	\centering
	\caption{The quantitative results for multi-dimensional data
		recovery by different methods (random missing cases). The \textbf{best} and \underline{second-best} values are highlighted.}
	\label{result1}
	\begin{tabular}{ccccccccccccccc}
		\toprule
		\multicolumn{2}{ c }{Method}&\multicolumn{2}{ c }{TNN \cite{zhang2014novel}}&\multicolumn{2}{ c }{TRLRF \cite{yuan2019tensor}}&\multicolumn{2}{ c }{FTNN \cite{jiang2020framelet}}&\multicolumn{2}{ c }{FCTN \cite{zheng2021fully}}&\multicolumn{2}{ c }{HLRTF         \cite{luo2022hlrtf}}&\multicolumn{2}{ c }{DTR}\\
		\midrule
		Data&Sampling rate&PSNR&SSIM&PSNR&SSIM&PSNR&SSIM&PSNR&SSIM&PSNR&SSIM&PSNR&SSIM\\
		\midrule
		HSI&0.1&33.56&0.9176&25.86&0.6817&37.71&0.9634&38.72&0.9487&\underline{40.88}&\underline{0.9839}&\textbf{45.05}&\textbf{0.9932}\\
		\emph{WDC}&0.2&38.20&0.9668&26.49&0.7134&\underline{43.87}&0.9879&43.59&0.9893&43.54&\underline{0.9906}&\textbf{47.08}&\textbf{0.9954}\\
		(256$\times$256$\times$191)&0.3&41.47&0.9827&28.72&0.7971&\underline{47.60}&0.9938&44.72&0.9916&45.81&\underline{0.9941}&\textbf{47.86}&\textbf{0.9959}\\
		\midrule
		HSI&0.1&32.70&0.8989&30.74&0.8439&38.00&0.9705&39.05&0.9692&\underline{41.71}&\underline{0.9845}&\textbf{45.87}&\textbf{0.9942}\\
		\emph{Pavia}&0.2&37.86&0.9610&32.05&0.8771&44.78&0.9921&46.74&0.9941&\underline{46.75}&\underline{0.9948}&\textbf{49.32}&\textbf{0.9968}\\
		(256$\times$256$\times$80)&0.3&41.57&0.9809&34.14&0.9170&48.80&0.9963&48.28&0.9958&\underline{49.09}&\underline{0.9968}&\textbf{50.22}&\textbf{0.9974}\\
		\midrule
		MSI&0.1&23.41&0.6456&23.13&0.5738&25.36&0.7818&26.26&0.7101&\underline{30.41}&\underline{0.8983}&\textbf{33.38}&\textbf{0.9582}\\
		\emph{Beads}&0.2&28.05&0.8330&26.63&0.7316&30.96&0.9196&31.84&0.8906&\underline{36.88}&\underline{0.9736}&\textbf{39.03}&\textbf{0.9872}\\
		(256$\times$256$\times$31)&0.3&31.82&0.9165&27.68&0.7742&35.17&0.9647&38.01&0.9671&\underline{41.30}&\underline{0.9905}&\textbf{43.89}&\textbf{0.9950}\\
		\midrule
		MSI&0.1&25.17&0.7853&24.34&0.7225&28.45&0.8995&27.39&0.8455&\underline{31.54}&\underline{0.9430}&\textbf{32.34}&\textbf{0.9643}\\
		\emph{Cloth}&0.2&29.73&0.9110&27.69&0.8473&33.54&0.9666&33.51&0.9569&\underline{38.12}&\underline{0.9861}&\textbf{39.76}&\textbf{0.9921}\\
		(256$\times$256$\times$31)&0.3&33.36&0.9577&28.72&0.8733&37.31&0.9855&36.72&0.9667&\underline{42.56}&\underline{0.9947}&\textbf{43.26}&\textbf{0.9957}\\
		\midrule
		MSI&0.1&31.94&0.7925&21.75&0.4215&35.96&0.8821&32.46&0.7666&\underline{40.45}&\underline{0.9362}&\textbf{40.81}&\textbf{0.9445}\\
		\emph{Flowers}&0.2&35.97&0.8660&26.01&0.5419&40.60&0.9230&38.85&0.8753&\underline{44.96}&\underline{0.9521}&\textbf{45.95}&\textbf{0.9698}\\
		(256$\times$256$\times$31)&0.3&38.66&0.8948&27.00&0.6326&43.53&0.9377&39.94&0.8922&\underline{48.33}&\underline{0.9714}&\textbf{48.34}&\textbf{0.9728}\\
		\midrule
		MSI&0.1&30.01&0.9113&31.01&0.8662&31.40&\textbf{0.9657}&29.23&0.8713&\underline{31.91}&0.9596&\textbf{33.95}&\underline{0.9615}\\
		\emph{Cd}&0.2&33.81&0.9537&34.35&0.9316&35.85&\textbf{0.9843}&36.58&0.9510&\underline{36.55}&0.9836&\textbf{39.17}&\underline{0.9839}\\
		(256$\times$256$\times$31)&0.3&36.93&0.9729&35.98&0.9491&\underline{38.78}&\textbf{0.9901}&38.39&0.9640&38.61&0.9884&\textbf{41.46}&\underline{0.9887}\\
		\midrule
		MSI&0.1&37.88&0.9679&37.94&0.9650&41.54&0.9871&40.65&0.9702&\underline{43.80}&\textbf{0.9890}&\textbf{43.83}&\underline{0.9882}\\
		\emph{Beers}&0.2&43.50&0.9902&40.75&0.9782&47.13&\underline{0.9957}&44.84&0.9878&\underline{47.93}&0.9948&\textbf{48.71}&\textbf{0.9958}\\
		(256$\times$256$\times$31)&0.3&47.48&0.9959&42.61&0.9841&50.58&\textbf{0.9979}&48.02&0.9946&\underline{50.79}&0.9968&\textbf{50.82}&\underline{0.9970}\\
		\midrule
		MSI&0.1&30.81&0.8622&29.57&0.7451&34.51&0.9482&33.37&0.8876&\underline{38.51}&\underline{0.9721}&\textbf{39.39}&\textbf{0.9771}\\
		\emph{Feathers}&0.2&35.52&0.9402&34.84&0.8995&39.62&0.9804&37.83&0.9426&\underline{43.94}&\underline{0.9892}&\textbf{44.12}&\textbf{0.9895}\\
		(256$\times$256$\times$31)&0.3&38.98&0.9692&35.20&0.9175&43.17&0.9896&38.95&0.9538&\textbf{47.58}&\textbf{0.9939}&\underline{46.35}&\underline{0.9902}\\
		\midrule
		MSI&0.1&38.62&0.8764&27.86&0.5878&40.54&0.8911&35.40&0.8297&\textbf{44.57}&\underline{0.9540}&\underline{44.34}&\textbf{0.9746}\\
		\emph{Statue}&0.2&41.75&0.9030&28.83&0.5992&43.65&0.9130&43.65&0.9188&\underline{47.73}&\underline{0.9649}&\textbf{48.81}&\textbf{0.9777}\\
		(256$\times$256$\times$31)&0.3&43.31&0.9115&30.53&0.6757&45.12&0.9200&44.51&0.9245&\underline{49.59}&\underline{0.9709}&\textbf{51.10}&\textbf{0.9825}\\
		\midrule
		MSI&0.1&31.27&0.8857&29.00&0.7253&35.67&0.9558&35.87&0.9272&\underline{37.90}&\underline{0.9624}&\textbf{38.86}&\textbf{0.9700}\\
		\emph{Toys }&0.2&36.22&0.9520&28.85&0.8235&41.12&0.9835&39.23&0.9577&\underline{45.34}&\underline{0.9894}&\textbf{45.89}&\textbf{0.9902}\\
		(256$\times$256$\times$31)&0.3&39.67&0.9748&36.36&0.9333&44.70&0.9908&40.30&0.9669&\textbf{47.80}&\textbf{0.9922}&\underline{47.05}&\underline{0.9912}\\
		\midrule
		MSI&0.1&27.40&0.7546&28.12&0.7211&\underline{31.03}&0.8924&30.02&0.7924&30.41&\underline{0.8983}&\textbf{34.89}&\textbf{0.9542}\\
		\emph{Jelly }&0.2&31.94&0.8850&30.61&0.8143&36.56&0.9650&31.78&0.8500&\underline{36.88}&\underline{0.9736}&\textbf{41.41}&\textbf{0.9868}\\
		(256$\times$256$\times$31)&0.3&35.35&0.9399&31.56&0.8451&40.28&0.9835&32.87&0.8780&\underline{41.30}&\underline{0.9905}&\textbf{43.82}&\textbf{0.9914}\\
		\midrule
		MSI&0.1&34.18&0.8950&32.14&0.8544&36.49&0.9356&36.92&0.9235&\underline{38.94}&\underline{0.9558}&\textbf{39.92}&\textbf{0.9644}\\
		\emph{Painting }&0.2&38.48&0.9531&33.38&0.9138&40.84&0.9718&39.04&0.9488&\underline{43.96}&\underline{0.9835}&\textbf{44.89}&\textbf{0.9847}\\
		(256$\times$256$\times$31)&0.3&41.88&0.9744&34.42&0.9333&44.33&0.9844&40.04&0.9569&\underline{47.23}&\underline{0.9911}&\textbf{47.90}&\textbf{0.9927}\\
		\midrule
		MSI&0.1&31.49&0.9043&30.10&0.8890&34.90&0.9574&34.88&0.9458&\underline{37.88}&\textbf{0.9730}&\textbf{38.20}&\underline{0.9721}\\
		\emph{Watercolors }&0.2&36.41&0.9652&34.08&0.9396&40.21&0.9859&37.04&0.9633&\underline{43.11}&\underline{0.9900}&\textbf{44.78}&\textbf{0.9925}\\
		(256$\times$256$\times$31)&0.3&39.88&0.9830&35.84&0.9552&43.79&0.9931&37.98&0.9700&\underline{46.53}&\underline{0.9948}&\textbf{46.78}&\textbf{0.9950}\\
		\midrule
		Video&0.1&22.81&0.8097&20.83&0.7368&22.24&0.7680&24.57&0.8525&\underline{25.06}&\underline{0.8693}&\textbf{26.85}&\textbf{0.9315}\\
		\emph{Bird}&0.2&26.38&0.9011&24.43&0.8583&26.49&0.9015&26.22&0.8924&\underline{29.86}&\underline{0.9422}&\textbf{31.84}&\textbf{0.9729}\\
		(288$\times$352$\times$3$\times$10)&0.3&29.16&0.9419&27.01&0.9088&29.80&0.9503&27.11&0.9098&\underline{33.55}&\underline{0.9720}&\textbf{34.82}&\textbf{0.9846}\\
		\midrule
		Video&0.1&21.49&0.8448&22.13&0.8529&20.22&0.8399&24.21&0.9103&\underline{24.68}&\underline{0.9380}&\textbf{26.13}&\textbf{0.9595}\\
		\emph{Sunflower}&0.2&25.74&0.9451&23.83&0.8705&23.81&0.9187&26.57&0.9329&\underline{27.37}&\underline{0.9577}&\textbf{29.30}&\textbf{0.9807}\\
		(288$\times$352$\times$3$\times$10)&0.3&28.21&0.9594&25.38&0.8767&26.42&0.9245&27.46&0.9390&\underline{29.68}&\underline{0.9708}&\textbf{32.18}&\textbf{0.9856}\\
		\midrule
		&0.1&30.18&0.8501&27.63&0.7458&32.93&0.9092&32.60&0.8767&\underline{35.91}&\underline{0.9478}&\textbf{37.59}&\textbf{0.9672}\\
		\emph{Average}&0.2&34.64&0.9284&30.19&0.8227&37.94&0.9593&37.15&0.9368&\underline{40.86}&\underline{0.9777}&\textbf{42.67}&\textbf{0.9864}\\
		&0.3&37.85&0.9570&32.08&0.8649&41.29&0.9735&38.89&0.9514&\underline{43.98}&\underline{0.9873}&\textbf{45.06}&\textbf{0.9904}\\
		\bottomrule
	\end{tabular}
\end{table*}
\begin{figure*}[!ht]
	\centering
	\subfloat{
		\begin{minipage}[b]{0.121\linewidth}
			\captionsetup{
				labelfont=bf,        
				textfont={rm}, 
				singlelinecheck=true 
			}
			\includegraphics[width=1\linewidth]{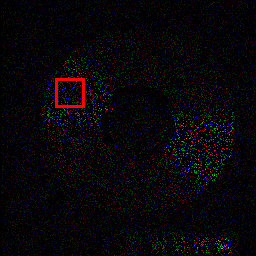}\vspace{1pt}
			\put(-25.1,0){\includegraphics[scale=1]{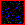}}\\
			\includegraphics[width=1\linewidth]{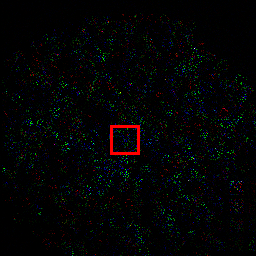}\vspace{1pt}
			\put(-25.1,0){\includegraphics[scale=1]{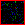}}\\
			\includegraphics[width=1\linewidth]{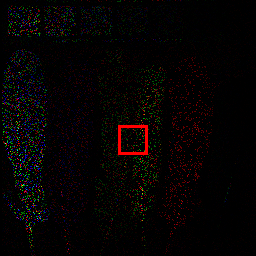}\vspace{1pt}
			\put(-25.1,0){\includegraphics[scale=1]{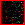}}\\
			\includegraphics[width=1\linewidth]{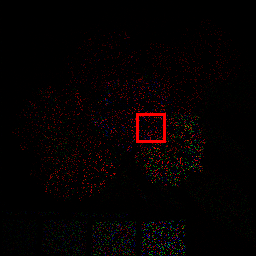}\vspace{1pt}
			\put(-25.1,0){\includegraphics[scale=1]{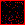}}
			\caption*{Observed}
	\end{minipage}}
	\hspace{-1.7mm}
	\subfloat{
		\begin{minipage}[b]{0.121\linewidth}
			\captionsetup{
				labelfont=bf,        
				textfont={rm}, 
				singlelinecheck=true 
			}
			\includegraphics[width=1\linewidth]{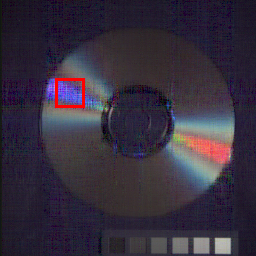}\vspace{1pt}
			\put(-25.1,0){\includegraphics[scale=1]{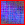}}\\
			\includegraphics[width=1\linewidth]{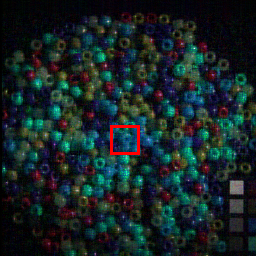}\vspace{1pt}
			\put(-25.1,0){\includegraphics[scale=1]{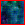}}\\
			\includegraphics[width=1\linewidth]{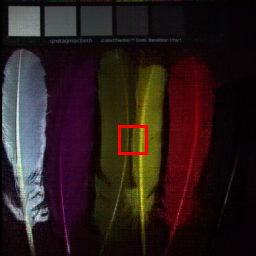}\vspace{1pt}
			\put(-25.1,0){\includegraphics[scale=1]{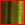}}\\
			\includegraphics[width=1\textwidth]{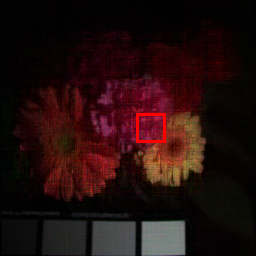}\vspace{1pt}
			\put(-25.1,0){\includegraphics[scale=1]{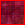}}
			\caption*{\fontfamily{Times New Roman}{TNN}}
	\end{minipage}}
	\hspace{-1.7mm}
	\subfloat{
		\begin{minipage}[b]{0.121\linewidth}
			\captionsetup{
				labelfont=bf,        
				textfont={rm}, 
				singlelinecheck=true 
			}
			\includegraphics[width=1\linewidth]{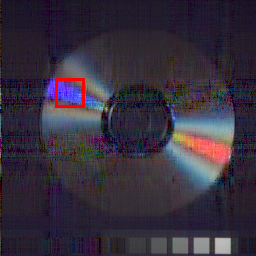}\vspace{1pt}
			\put(-25.1,0){\includegraphics[scale=1]{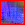}}\\
			\includegraphics[width=1\linewidth]{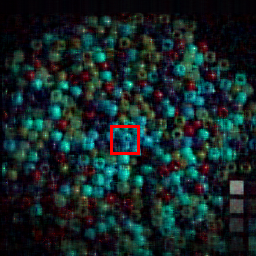}\vspace{1pt}
			\put(-25.1,0){\includegraphics[scale=1]{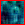}}\\
			\includegraphics[width=1\linewidth]{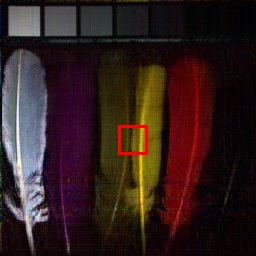}\vspace{1pt}
			\put(-25.1,0){\includegraphics[scale=1]{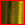}}\\
			\includegraphics[width=1\linewidth]{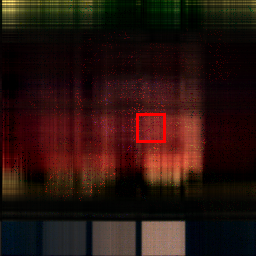}\vspace{1pt}
			\put(-25.1,0){\includegraphics[scale=1]{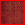}}
			\caption*{\fontfamily{Times New Roman}{TRLRF}}
	\end{minipage}}
	\hspace{-1.7mm}
	\subfloat{
		\begin{minipage}[b]{0.121\linewidth}
			\captionsetup{
				labelfont=bf,        
				textfont={rm}, 
				singlelinecheck=true 
			}
			\includegraphics[width=1\linewidth]{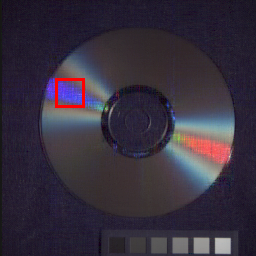}\vspace{1pt}
			\put(-25.1,0){\includegraphics[scale=1]{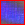}}\\
			\includegraphics[width=1\linewidth]{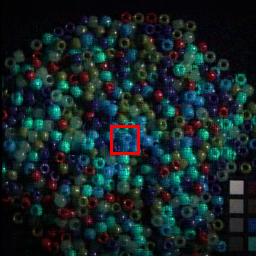}\vspace{1pt}
			\put(-25.1,0){\includegraphics[scale=1]{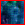}}\\
			\includegraphics[width=1\linewidth]{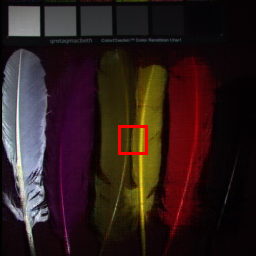}\vspace{1pt}
			\put(-25.1,0){\includegraphics[scale=1]{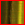}}\\
			\includegraphics[width=1\linewidth]{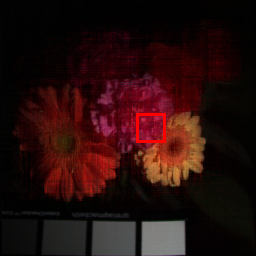}\vspace{1pt}
			\put(-25.1,0){\includegraphics[scale=1]{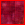}}
			\caption*{\fontfamily{Times New Roman}{FTNN}}
	\end{minipage}}
	\hspace{-1.7mm}
	\subfloat{
		\begin{minipage}[b]{0.121\linewidth}
			\captionsetup{
				labelfont=bf,        
				textfont={rm}, 
				singlelinecheck=true 
			}
			\includegraphics[width=1\linewidth]{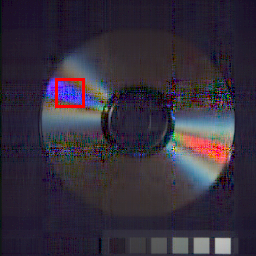}\vspace{1pt}
			\put(-25.1,0){\includegraphics[scale=1]{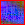}}\\
			\includegraphics[width=1\linewidth]{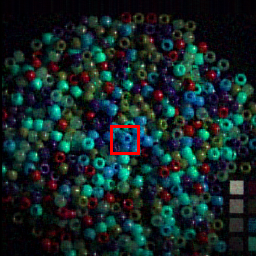}\vspace{1pt}
			\put(-25.1,0){\includegraphics[scale=1]{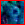}}\\
			\includegraphics[width=1\linewidth]{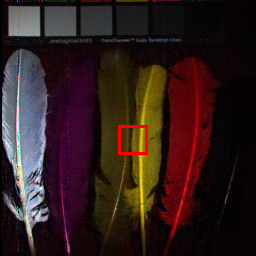}\vspace{1pt}
			\put(-25.1,0){\includegraphics[scale=1]{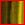}}\\
			\includegraphics[width=1\linewidth]{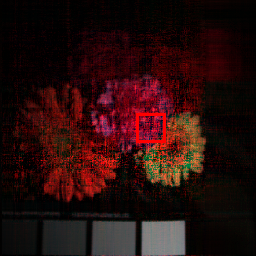}\vspace{1pt}
			\put(-25.1,0){\includegraphics[scale=1]{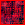}}
			\caption*{\fontfamily{Times New Roman}{FCTN}}
	\end{minipage}}
	\hspace{-1.7mm}
	\subfloat{
		\begin{minipage}[b]{0.121\linewidth}
			\captionsetup{
				labelfont=bf,        
				textfont={rm}, 
				singlelinecheck=true 
			}
			\includegraphics[width=1\linewidth]{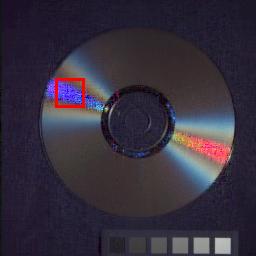}\vspace{1pt}
			\put(-25.1,0){\includegraphics[scale=1]{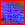}}\\
			\includegraphics[width=1\linewidth]{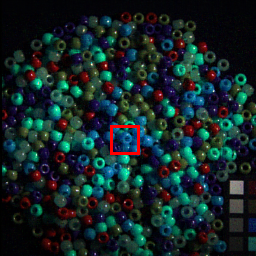}\vspace{1pt}
			\put(-25.1,0){\includegraphics[scale=1]{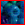}}\\
			\includegraphics[width=1\linewidth]{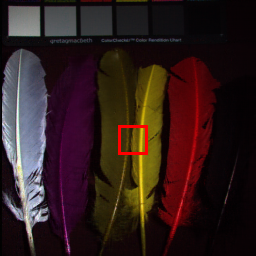}\vspace{1pt}
			\put(-25.1,0){\includegraphics[scale=1]{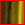}}\\
			\includegraphics[width=1\linewidth]{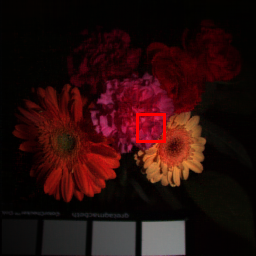}\vspace{1pt}
			\put(-25.1,0){\includegraphics[scale=1]{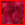}}
			\caption*{\fontfamily{Times New Roman}{HLRTF}}
	\end{minipage}}
	\hspace{-1.7mm}
	\subfloat{
		\begin{minipage}[b]{0.121\linewidth}
			\captionsetup{
				labelfont=bf,        
				textfont={rm}, 
				singlelinecheck=true 
			}
			\includegraphics[width=1\linewidth]{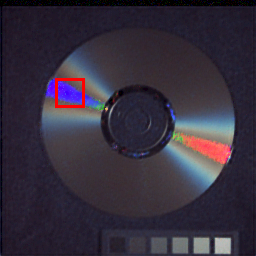}\vspace{1pt}
			\put(-25.1,0){\includegraphics[scale=1]{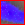}}\\
			\includegraphics[width=1\linewidth]{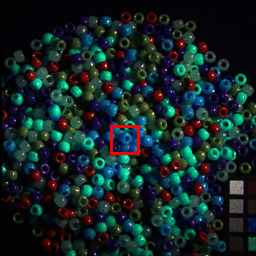}\vspace{1pt}
			\put(-25.1,0){\includegraphics[scale=1]{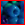}}\\
			\includegraphics[width=1\linewidth]{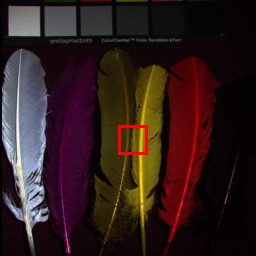}\vspace{1pt}
			\put(-25.1,0){\includegraphics[scale=1]{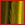}}\\
			\includegraphics[width=1\linewidth]{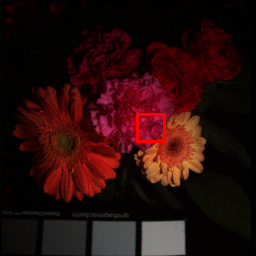}\vspace{1pt}
			\put(-25.1,0){\includegraphics[scale=1]{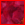}}
			
			\caption*{\fontfamily{Times New Roman}{DTR}}
	\end{minipage}}
	\hspace{-1.7mm}
	\subfloat{
		\begin{minipage}[b]{0.121\linewidth}
			\captionsetup{
				labelfont=bf,        
				textfont={rm}, 
				singlelinecheck=true 
			}
			\includegraphics[width=1\linewidth]{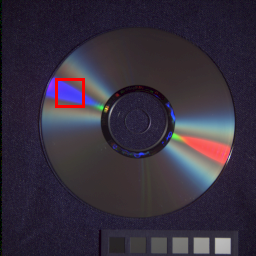}\vspace{1pt}
			\put(-25.1,0){\includegraphics[scale=1]{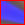}}\\
			\includegraphics[width=1\linewidth]{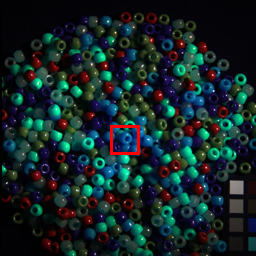}\vspace{1pt}
			\put(-25.1,0){\includegraphics[scale=1]{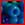}}\\
			\includegraphics[width=1\linewidth]{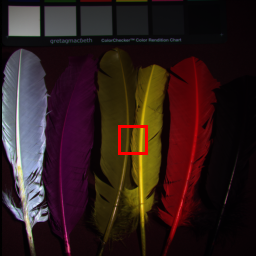}\vspace{1pt}
			\put(-25.1,0){\includegraphics[scale=1]{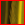}}\\
			\includegraphics[width=1\linewidth]{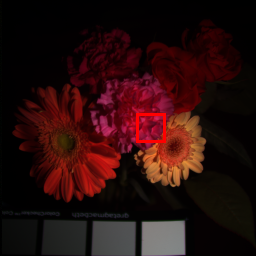}\vspace{1pt}
			\put(-25.1,0){\includegraphics[scale=1]{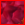}}
			\caption*{\fontfamily{Times New Roman}{Original}}
	\end{minipage}}
	\caption{The recovered pseudo-color images by different methods. From top to bottom: MSI\ \emph{Cd} (R:10, G:20, B:30), \emph{Beads} (R:10, G:20, B:30), \emph{Feathers} (R:31, G:15, B:4), and \emph{Flower} (R:31, G:15, B:4) with the random missing (SR = 0.1).}	
	\label{visual2}
\end{figure*}

\begin{figure*}[!ht]
	\centering
	\subfloat{
		\begin{minipage}[b]{0.121\linewidth}
			\captionsetup{
				labelfont=bf,        
				textfont={rm}, 
				singlelinecheck=true 
			}
			\includegraphics[width=1\linewidth]{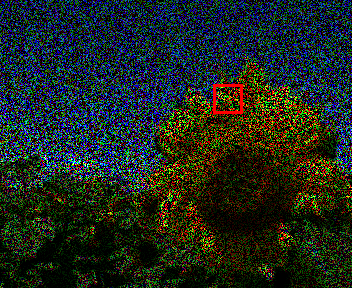}\vspace{1pt}
			\put(-25.1,0){\includegraphics[scale=1]{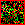}}\\
			\includegraphics[width=1\linewidth]{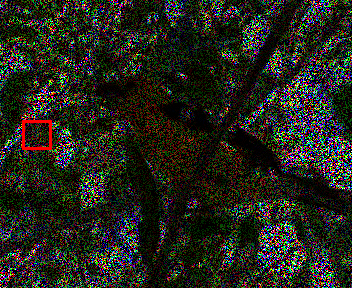}\vspace{1pt}
			\put(-25.1,0){\includegraphics[scale=1]{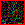}}
			
			\caption*{\fontfamily{Times New Roman}{Observed}}
	\end{minipage}}
	\hspace{-1.7mm}
	\subfloat{
		\begin{minipage}[b]{0.121\linewidth}
			\captionsetup{
				labelfont=bf,        
				textfont={rm}, 
				singlelinecheck=true 
			}
		    \includegraphics[width=1\linewidth]{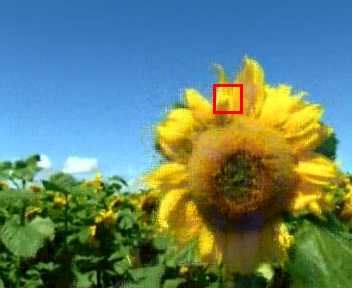}\vspace{1pt}
		    \put(-25.1,0){\includegraphics[scale=1]{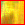}}\\
			\includegraphics[width=1\linewidth]{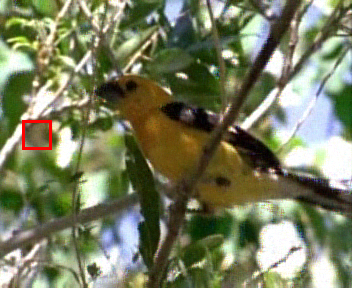}\vspace{1pt}
			\put(-25.1,0){\includegraphics[scale=1]{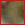}}
			\caption*{\fontfamily{Times New Roman}{TNN}}
	\end{minipage}}
	\hspace{-1.7mm}
	\subfloat{
		\begin{minipage}[b]{0.121\linewidth}
			\captionsetup{
				labelfont=bf,        
				textfont={rm}, 
				singlelinecheck=true 
			}
			\includegraphics[width=1\linewidth]{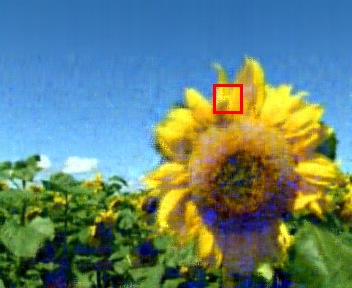}\vspace{1pt}
			\put(-25.1,0){\includegraphics[scale=1]{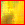}}\\
			\includegraphics[width=1\linewidth]{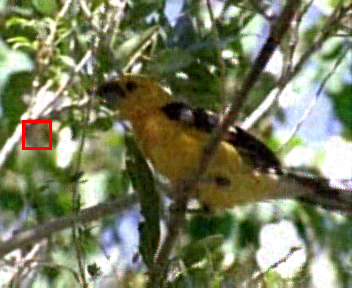}\vspace{1pt}
			\put(-25.1,0){\includegraphics[scale=1]{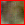}}
			
			\caption*{\fontfamily{Times New Roman}{TRLRF}}
	\end{minipage}}
	\hspace{-1.7mm}
	\subfloat{
		\begin{minipage}[b]{0.121\linewidth}
			\captionsetup{
				labelfont=bf,        
				textfont={rm}, 
				singlelinecheck=true 
			}
			\includegraphics[width=1\linewidth]{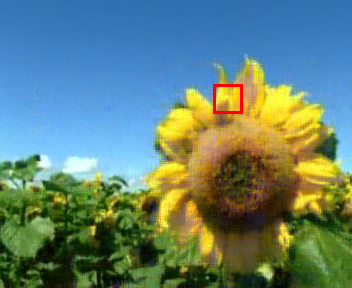}\vspace{1pt}
			\put(-25.1,0){\includegraphics[scale=1]{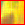}}\\
			\includegraphics[width=1\linewidth]{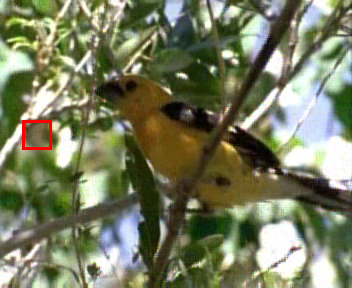}\vspace{1pt}
			\put(-25.1,0){\includegraphics[scale=1]{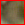}}
			\caption*{\fontfamily{Times New Roman}{FTNN}}
	\end{minipage}}
	\hspace{-1.7mm}
	\subfloat{
		\begin{minipage}[b]{0.121\linewidth}
			\captionsetup{
				labelfont=bf,        
				textfont={rm}, 
				singlelinecheck=true 
			}
			\includegraphics[width=1\linewidth]{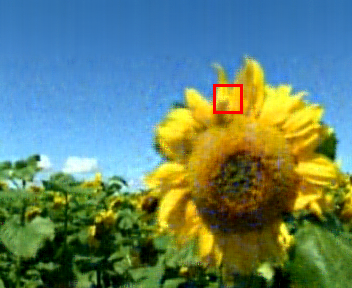}\vspace{1pt}
			\put(-25.1,0){\includegraphics[scale=1]{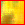}}\\
			\includegraphics[width=1\linewidth]{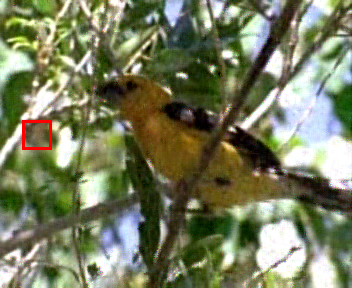}\vspace{1pt}
			\put(-25.1,0){\includegraphics[scale=1]{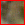}}
			
			\caption*{\fontfamily{Times New Roman}{FCTN}}
	\end{minipage}}
	\hspace{-1.7mm}
	\subfloat{
		\begin{minipage}[b]{0.121\linewidth}
			\captionsetup{
				labelfont=bf,        
				textfont={rm}, 
				singlelinecheck=true 
			}
			\includegraphics[width=1\linewidth]{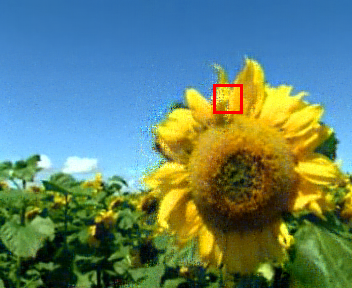}\vspace{1pt}
			\put(-25.1,0){\includegraphics[scale=1]{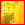}}\\
			\includegraphics[width=1\linewidth]{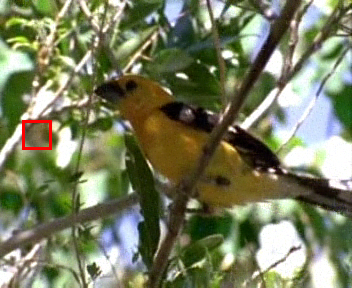}\vspace{1pt}
			\put(-25.1,0){\includegraphics[scale=1]{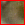}}
			
			\caption*{\fontfamily{Times New Roman}{HLRTF}}
	\end{minipage}}
	\hspace{-1.7mm}
	\subfloat{
		\begin{minipage}[b]{0.121\linewidth}
			\captionsetup{
				labelfont=bf,        
				textfont={rm}, 
				singlelinecheck=true 
			}
			\includegraphics[width=1\linewidth]{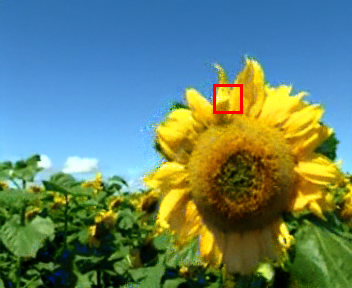}\vspace{1pt}
			\put(-25.1,0){\includegraphics[scale=1]{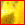}}\\
			\includegraphics[width=1\linewidth]{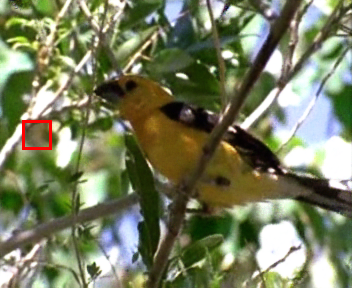}\vspace{1pt}
			\put(-25.1,0){\includegraphics[scale=1]{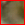}}
			\caption*{\fontfamily{Times New Roman}{DTR}}
	\end{minipage}}
	\hspace{-1.7mm}
	\subfloat{
		\begin{minipage}[b]{0.121\linewidth}
			\captionsetup{
				labelfont=bf,        
				textfont={rm}, 
				singlelinecheck=true 
			}
			\includegraphics[width=1\linewidth]{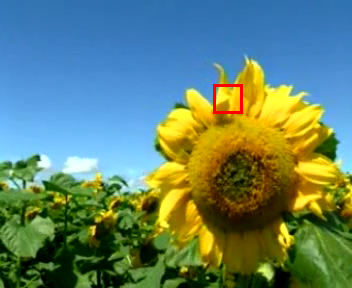}\vspace{1pt}
			\put(-25.1,0){\includegraphics[scale=1]{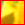}}\\
			\includegraphics[width=1\linewidth]{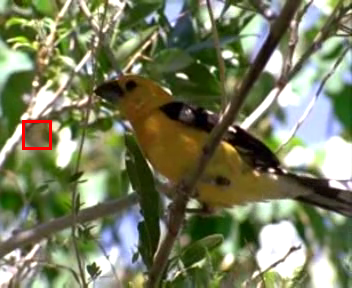}\vspace{1pt}
			\put(-25.1,0){\includegraphics[scale=1]{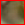}}
			\caption*{\fontfamily{Times New Roman}{Original}}
	\end{minipage}}
	\caption{The 10-th frame of the recovered results by all comparative methods on videos \emph{Sunflower} and \emph{Bird} with the random missing (SR = 0.3).}	
	\label{visual3}
\end{figure*}
	\section{Experiments}
	\label{section:5}
	In this section, we conduct a series of experiments to verify the effectiveness and superiority of the proposed DTR. We first introduce some experimental settings in Section \ref{Settings}, and then present experimental results in Section \ref{Results}.
	
	\subsection{Experimental Settings}\label{Settings}
	$\mathbf{Compared\ Methods.}$
	To comprehensively evaluate the effectiveness of DTR, we compare it with five classic multi-dimensional
	data recovery methods, i.e., TNN \cite{zhang2014novel}, TRLRF \cite{yuan2019tensor}, FTNN \cite{jiang2020framelet}, FCTN \cite{zheng2021fully}, and HLRTF \cite{luo2022hlrtf}. These methods are all tensor decomposition-based methods. Note that TNN and HLRTF adopt shallow SVD and NMF to characterize the frontal slices of latent tensor, respectively. To ensure optimal performance, all
	parameters associated with the compared methods are carefully
	adjusted according to the suggestions provided by the respective
	authors in their articles.
	
	$\mathbf{Implementation.}$ We employ two HSIs (i.e., \emph{WDC}\footnote[2]{Available at https://engineering.purdue.edu/biehl/MultiSpec/\\hyperspectral.html
	} and \emph{Pavia}\footnote[3]{Available at https://www.ehu.eus/ccwintco/index.php/\\Hyperspectral\_Remote\_Sensing\_Scenes}), eleven MSIs\footnote[4]{Available at http://www.cs.columbia.edu/CAVE/databases/multispectral/} (i.e., \emph{Beads}, \emph{Cloth}, \emph{Flowers}, \emph{Cd}, \emph{Beers}, \emph{Feathers}, \emph{Statue}, \emph{Toys}, \emph{Jelly}, \emph{Painting}, and \emph{Watercolors}), and two videos\footnote[5]{Available at http://trace.eas.asu.edu/yuv/.} (i.e., \emph{Bird} and \emph{Sunflower}) to test the performance of the
	proposed DTR. The sampling rates for both random and tube missing cases are set as $\text{SR}=\{0.1, 0.2, 0.3\}$. Before the experiment, the gray values of all datasets are normalized band-by-band into the interval $[0,1]$. All experiments are performed
	on a PC equipped with two Intel(R) Xeon(R) Silver 4210R CPUs, two NVIDIA RTX 3090 GPUs, and 128GB RAM. The proposed DTR and HLRTF are implemented by using the Python and the PyTorch library with GPU calculation, and other comparison methods are implemented by MATLAB (R2018a) with CPU calculation. In our experiments,  $g_\theta(\cdot)$ is set as the U-Net, $f_\xi(\cdot)$ is set as the FCN, $\hat{n}_3$ value is set as the $n_3$, the layer number for both modules is set to 2, the Adam algorithm is adopted to train the model with learning rate 0.001, and the maximum iteration step for both HLRTF and the proposed DTR is set to 7000.
	\raggedbottom
	
	$\mathbf{Evaluation\ Metrics.}$
	Two quantitative evaluation metrics peak-signal-to-noise-ratio (PSNR) and structural similarity (SSIM) are selected to evaluate the overall quality
	of the recovered results. In general, higher PSNR
	and SSIM values indicate better results.

	\begin{table*}[t]\scriptsize
		\centering
		\caption{The quantitative results for multi-dimensional data
			recovery by different methods (tube missing cases). The \textbf{best} and \underline{second-best} values are highlighted.}
		\label{result2}
		\begin{tabular}{ccccccccccccccc}
			\toprule
			\multicolumn{2}{ c }{Method}&\multicolumn{2}{ c }{TNN \cite{zhang2014novel}}&\multicolumn{2}{ c }{TRLRF \cite{yuan2019tensor}}&\multicolumn{2}{ c }{FTNN  \cite{jiang2020framelet}}&\multicolumn{2}{ c }{FCTN \cite{zheng2021fully}}&\multicolumn{2}{ c }{HLRTF \cite{luo2022hlrtf}}&\multicolumn{2}{ c }{DTR}\\
			\midrule
			Data&Sampling rate&PSNR&SSIM&PSNR&SSIM&PSNR&SSIM&PSNR&SSIM&PSNR&SSIM&PSNR&SSIM\\
			\midrule
			HSI&0.1&20.55&0.3704&19.35&0.3302&20.62&0.4144&19.16&0.3530&\textbf{22.15}&\textbf{0.4993}&\underline{21.97}&\underline{0.4905}\\
			\emph{WDC}&0.2&22.48&0.5003&21.79&0.4922&22.99&0.5521&21.78&0.4933&\underline{24.05}&\underline{0.6011}&\textbf{24.26}&\textbf{0.6244}\\
			(256$\times$256$\times$191)&0.3&24.17&0.6035&22.66&0.5554&24.82&0.6544&22.63&0.5557&\underline{25.79}&\underline{0.6916}&\textbf{26.15}&\textbf{0.7265}\\
			\midrule
			HSI&0.1&21.01&0.3334&19.31&0.2798&20.73&0.3679&19.50&0.3212&\underline{22.60}&\underline{0.4791}&\textbf{24.83}&\textbf{0.6417}\\
			\emph{Pavia}&0.2&22.94&0.4614&21.68&0.4263&23.31&0.5101&22.32&0.4595&\underline{24.39}&\underline{0.5737}&\textbf{27.48}&\textbf{0.7696}\\
			(256$\times$256$\times$80)&0.3&24.56&0.5777&23.14&0.5262&25.11&0.6281&23.13&0.5259&\underline{26.23}&\underline{0.6860}&\textbf{29.99}&\textbf{0.8595}\\
			\midrule
			MSI&0.1&17.34&0.2868&16.29&0.2440&17.52&0.3712&15.35&0.2571&\underline{19.16}&\underline{0.4555}&\textbf{20.61}&\textbf{0.6226}\\
			\emph{Beads}&0.2&19.27&0.4248&18.02&0.3674&20.12&0.5281&18.22&0.3787&\underline{21.77}&\underline{0.6242}&\textbf{23.72}&\textbf{0.7696}\\
			(256$\times$256$\times$31)&0.3&21.28&0.5501&18.90&0.4376&22.38&\underline{0.6573}&19.05&0.4512&\underline{23.03}&0.6549&\textbf{26.78}&\textbf{0.8728}\\
			\midrule
			MSI&0.1&18.27&0.3961&17.00&0.3510&18.18&0.4428&15.72&0.3456&\textbf{19.85}&\underline{0.5025}&\underline{19.40}&\textbf{0.5148}\\
			\emph{Cloth}&0.2&19.84&0.4967&19.37&0.4885&20.05&0.5366&19.45&0.4880&\underline{21.01}&\underline{0.5737}&\textbf{21.51}&\textbf{0.6795}\\
			(256$\times$256$\times$31)&0.3&21.37&0.5917&20.42&0.5559&21.75&0.6287&20.44&0.5567&\underline{22.76}&\underline{0.6704}&\textbf{23.77}&\textbf{0.7848}\\
			\midrule
			MSI&0.1&24.75&0.5999&21.34&0.3660&21.51&0.6358&23.45&0.5638&\underline{27.78}&\underline{0.7611}&\textbf{29.36}&\textbf{0.8155}\\
			\emph{Flower}&0.2&27.95&0.6943&24.93&0.5120&27.93&0.7510&26.62&0.6399&\underline{31.25}&\underline{0.8473}&\textbf{32.49}&\textbf{0.8993}\\
			(256$\times$256$\times$31)&0.3&30.49&0.7686&25.53&0.5359&31.06&0.8233&27.57&0.6710&\underline{33.63}&\underline{0.9005}&\textbf{35.33}&\textbf{0.9182}\\
			\midrule
			MSI&0.1&24.84&0.7900&21.12&0.5117&22.00&0.8092&22.55&0.6321&\underline{26.77}&\underline{0.8888}&\textbf{29.26}&\textbf{0.8622}\\
			\emph{Cd}&0.2&28.35&0.8776&24.57&0.7237&27.81&0.9060&25.07&0.7772&\underline{30.62}&\textbf{0.9383}&\textbf{33.64}&\underline{0.9321}\\
			(256$\times$256$\times$31)&0.3&31.40&0.9214&25.52&0.7372&31.83&0.9462&25.79&0.7584&\underline{33.86}&\underline{0.9595}&\textbf{36.94}&\textbf{0.9617}\\
			\midrule
			MSI&0.1&25.43&0.8002&22.62&0.6764&18.03&0.5032&20.05&0.6876&\underline{29.34}&\underline{0.8965}&\textbf{30.65}&\textbf{0.8846}\\
			\emph{Beers}&0.2&31.29&0.9171&25.77&0.7864&25.94&0.8348&26.75&0.8202&\underline{35.24}&\textbf{0.9647}&\textbf{35.54}&\underline{0.9552}\\
			(256$\times$256$\times$31)&0.3&35.09&0.9569&26.69&0.8010&34.09&0.9561&27.60&0.8290&\underline{38.41}&\underline{0.9782}&\textbf{39.74}&\textbf{0.9783}\\
			\midrule
			MSI&0.1&21.99&0.6267&18.59&0.3422&18.06&0.6411&17.57&0.4196&\underline{23.88}&\textbf{0.7711}&\textbf{24.60}&\underline{0.7436}\\
			\emph{Feathers}&0.2&24.89&0.7375&22.30&0.5276&21.97&0.7459&22.76&0.5688&\underline{26.74}&\underline{0.8415}&\textbf{27.88}&\textbf{0.8722}\\
			(256$\times$256$\times$31)&0.3&27.07&0.8066&23.49&0.5808&26.44&0.8349&23.80&0.6210&\underline{29.61}&\underline{0.8979}&\textbf{31.61}&\textbf{0.9272}\\
			\midrule
			MSI&0.1&30.60&0.7611&18.14&0.1772&24.75&0.7420&27.23&0.6849&\textbf{33.26}&\underline{0.8527}&\underline{32.43}&\textbf{0.8853}\\
			\emph{Statue}&0.2&34.70&0.8310&22.99&0.4487&34.45&0.8397&31.41&0.7692&\underline{36.85}&\underline{0.9158}&\textbf{36.15}&\textbf{0.9383}\\
			(256$\times$256$\times$31)&0.3&36.20&0.8565&24.73&0.4876&36.63&0.8672&34.14&0.7918&\textbf{38.21}&\underline{0.9305}&\underline{38.05}&\textbf{0.9400}\\
			\midrule
			MSI&0.1&20.97&0.5969&20.23&0.4483&15.35&0.5736&18.15&0.5268&\underline{23.33}&\underline{0.7469}&\textbf{24.36}&\textbf{0.7532}\\
			\emph{Toys}&0.2&24.13&0.7139&21.89&0.5281&20.82&0.7020&23.05&0.6645&\underline{26.86}&\underline{0.8403}&\textbf{27.95}&\textbf{0.8706}\\
			(256$\times$256$\times$31)&0.3&26.75&0.7947&24.12&0.6930&25.27&0.8009&24.33&0.7127&\underline{29.54}&\underline{0.8890}&\textbf{30.02}&\textbf{0.9071}\\
			\midrule
			MSI&0.1&17.18&0.2835&14.85&0.1794&16.35&0.3747&13.70&0.1883&\underline{19.29}&\underline{0.4574}&\textbf{22.53}&\textbf{0.6810}\\
			\emph{Jelly}&0.2&19.60&0.4090&17.86&0.3293&18.91&0.4848&17.81&0.3295&\underline{22.05}&\underline{0.5953}&\textbf{25.11}&\textbf{0.8052}\\
			(256$\times$256$\times$31)&0.3&21.78&0.5421&18.96&0.4073&21.23&0.6017&18.97&0.4044&\underline{24.92}&\underline{0.7360}&\textbf{27.64}&\textbf{0.8752}\\
			\midrule
			MSI&0.1&25.19&0.6451&21.57&0.3818&21.71&0.6280&23.11&0.6029&\underline{27.62}&\underline{0.7580}&\textbf{29.35}&\textbf{0.7692}\\
			\emph{Painting}&0.2&28.19&0.7584&25.36&0.5776&26.51&0.7543&26.53&0.6827&\underline{30.82}&\textbf{0.8389}&\textbf{30.98}&\underline{0.8297}\\
			(256$\times$256$\times$31)&0.3&30.81&0.8265&27.40&0.7008&30.32&0.8367&27.29&0.7098&\underline{32.91}&\underline{0.8840}&\textbf{33.82}&\textbf{0.8858}\\
			\midrule
			MSI&0.1&22.51&0.5889&20.70&0.4980&16.35&0.4059&19.35&0.5236&\textbf{24.41}&\textbf{0.7097}&\underline{24.11}&\underline{0.6742}\\
			\emph{Watercolors}&0.2&24.74&0.7027&23.60&0.6551&21.91&0.5966&23.64&0.6577&\underline{26.92}&\underline{0.8060}&\textbf{27.34}&\textbf{0.8131}\\
			(256$\times$256$\times$31)&0.3&26.58&0.7819&24.61&0.7015&25.41&0.7466&24.66&0.7062&\underline{28.90}&\underline{0.8649}&\textbf{30.16}&\textbf{0.8841}\\
			\midrule
			Video&0.1&13.55&0.2178&12.78&0.3104&12.15&0.4171&13.10&0.3302&\underline{16.21}&\underline{0.6401}&\textbf{22.30}&\textbf{0.8523}\\
			\emph{Bird}&0.2&17.14&0.6250&14.60&0.4381&16.46&0.6271&14.59&0.4305&\underline{20.84}&\underline{0.7976}&\textbf{26.81}&\textbf{0.9337}\\
			(288$\times$352$\times$3$\times$10)&0.3&20.26&0.7640&19.01&0.6756&20.28&0.7810&15.42&0.4962&\underline{24.17}&\underline{0.8751}&\textbf{31.09}&\textbf{0.9702}\\
			\midrule
			Video&0.1&17.41&0.7897&14.75&0.6046&16.72&0.7683&16.78&0.7410&\underline{20.27}&\underline{0.8744}&\textbf{23.39}&\textbf{0.9353}\\
			\emph{Sunflower}&0.2&20.71&0.8823&17.79&0.7970&20.18&0.8667&17.62&0.7654&\underline{24.20}&\underline{0.9577}&\textbf{26.83}&\textbf{0.9647}\\
			(288$\times$352$\times$3$\times$10)&0.3&23.40&0.9248&21.66&0.8796&22.78&0.9140&18.22&0.7990&\underline{26.70}&\underline{0.9582}&\textbf{29.57}&\textbf{0.9776}\\
			\midrule
			&0.1&21.44&0.5391&18.58&0.3800&18.67&0.5397&18.98&0.4785&\underline{23.73}&\underline{0.6862}&\textbf{25.28}&\textbf{0.7417}\\
			\emph{Average}&0.2&24.41&0.6688&21.50&0.5399&23.29&0.6824&22.51&0.5950&\underline{26.91}&\underline{0.7811}&\textbf{28.51}&\textbf{0.8438}\\
			&0.3&26.75&0.7511&23.12&0.6184&26.63&0.7785&23.54&0.6393&\underline{29.24}&\underline{0.8384}&\textbf{31.38}&\textbf{0.8979}\\
			\bottomrule
		\end{tabular}
	\end{table*}
	\begin{figure*}[!ht]
		\centering
		\subfloat{
			\begin{minipage}[b]{0.121\linewidth}
				\captionsetup{
					labelfont=bf,        
					textfont={rm}, 
					singlelinecheck=true 
				}
				\includegraphics[width=1\linewidth]{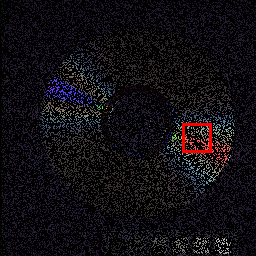}\vspace{1pt}
				\put(-25.1,0){\includegraphics[scale=1]{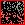}}\\
				\includegraphics[width=1\linewidth]{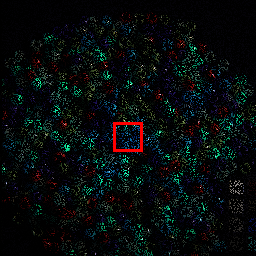}\vspace{1pt}
				\put(-25.1,0){\includegraphics[scale=1]{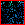}}\\
				\includegraphics[width=1\linewidth]{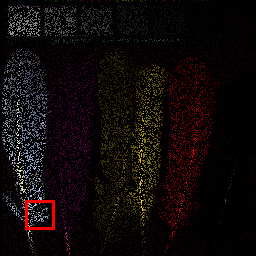}\vspace{1pt}
				\put(-25.1,0){\includegraphics[scale=1]{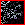}}\\
				\includegraphics[width=1\linewidth]{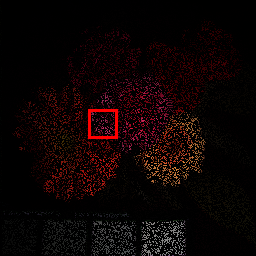}\vspace{1pt}
				\put(-25.1,0){\includegraphics[scale=1]{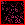}}
				\caption*{\fontfamily{Times New Roman}{Observed}}
		\end{minipage}}
		\hspace{-1.7mm}
		\subfloat{
			\begin{minipage}[b]{0.121\linewidth}
				\captionsetup{
					labelfont=bf,        
					textfont={rm}, 
					singlelinecheck=true 
				}
				\includegraphics[width=1\linewidth]{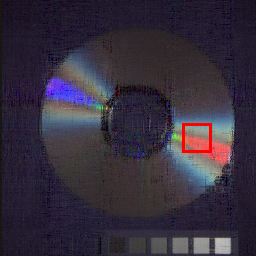}\vspace{1pt}
				\put(-25.1,0){\includegraphics[scale=1]{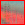}}\\
				\includegraphics[width=1\linewidth]{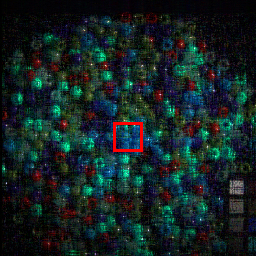}\vspace{1pt}
				\put(-25.1,0){\includegraphics[scale=1]{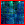}}\\
				\includegraphics[width=1\linewidth]{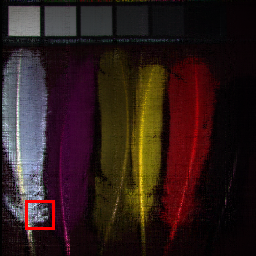}\vspace{1pt}
				\put(-25.1,0){\includegraphics[scale=1]{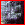}}\\
				\includegraphics[width=1\textwidth]{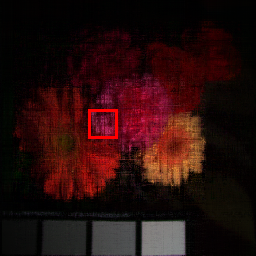}\vspace{1pt}
				\put(-25.1,0){\includegraphics[scale=1]{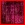}}
				\caption*{\fontfamily{Times New Roman}{TNN}}
		\end{minipage}}
		\hspace{-1.7mm}
		\subfloat{
			\begin{minipage}[b]{0.121\linewidth}
				\captionsetup{
					labelfont=bf,        
					textfont={rm}, 
					singlelinecheck=true 
				}
				\includegraphics[width=1\linewidth]{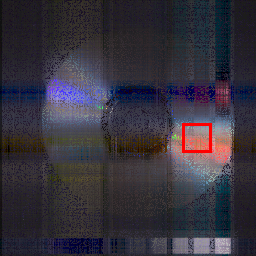}\vspace{1pt}
				\put(-25.1,0){\includegraphics[scale=1]{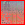}}\\
				\includegraphics[width=1\linewidth]{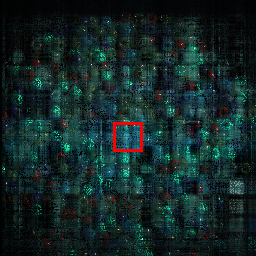}\vspace{1pt}
				\put(-25.1,0){\includegraphics[scale=1]{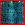}}\\
				\includegraphics[width=1\linewidth]{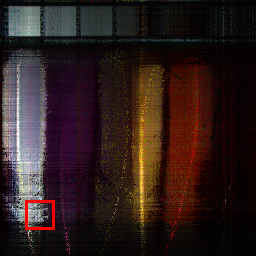}\vspace{1pt}
				\put(-25.1,0){\includegraphics[scale=1]{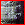}}\\
				\includegraphics[width=1\linewidth]{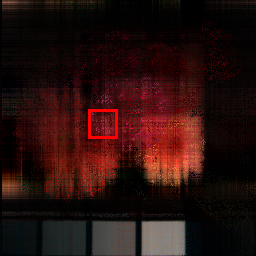}\vspace{1pt}
				\put(-25.1,0){\includegraphics[scale=1]{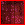}}
				\caption*{\fontfamily{Times New Roman}{TRLRF}}
		\end{minipage}}
		\hspace{-1.7mm}
		\subfloat{
			\begin{minipage}[b]{0.121\linewidth}
				\captionsetup{
					labelfont=bf,        
					textfont={rm}, 
					singlelinecheck=true 
				}
				\includegraphics[width=1\linewidth]{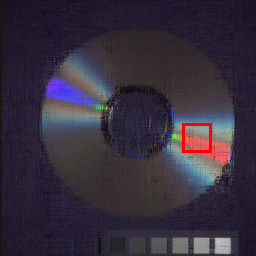}\vspace{1pt}
				\put(-25.1,0){\includegraphics[scale=1]{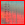}}\\
				\includegraphics[width=1\linewidth]{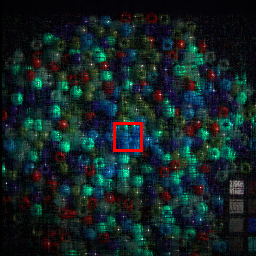}\vspace{1pt}
				\put(-25.1,0){\includegraphics[scale=1]{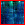}}\\
				\includegraphics[width=1\linewidth]{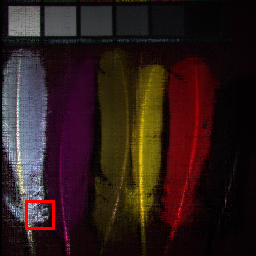}\vspace{1pt}
				\put(-25.1,0){\includegraphics[scale=1]{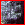}}\\
				\includegraphics[width=1\linewidth]{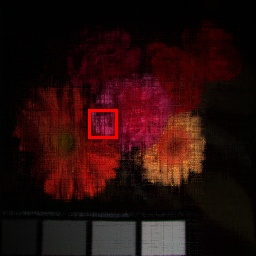}\vspace{1pt}
				\put(-25.1,0){\includegraphics[scale=1]{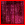}}
				\caption*{\fontfamily{Times New Roman}{FTNN}}
		\end{minipage}}
		\hspace{-1.7mm}
		\subfloat{
			\begin{minipage}[b]{0.121\linewidth}
				\captionsetup{
					labelfont=bf,        
					textfont={rm}, 
					singlelinecheck=true 
				}
				\includegraphics[width=1\linewidth]{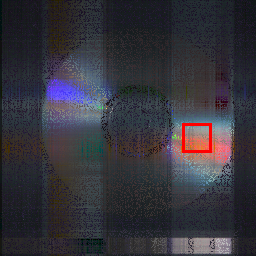}\vspace{1pt}
				\put(-25.1,0){\includegraphics[scale=1]{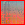}}\\
				\includegraphics[width=1\linewidth]{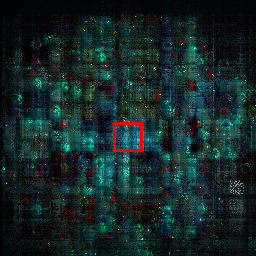}\vspace{1pt}
				\put(-25.1,0){\includegraphics[scale=1]{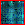}}\\
				\includegraphics[width=1\linewidth]{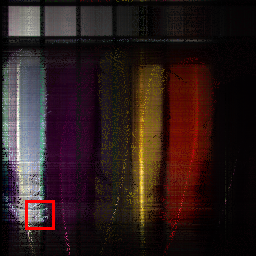}\vspace{1pt}
				\put(-25.1,0){\includegraphics[scale=1]{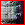}}\\
				\includegraphics[width=1\linewidth]{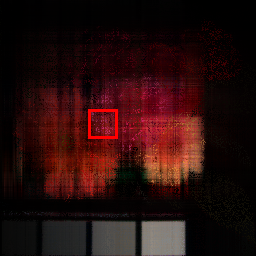}\vspace{1pt}
				\put(-25.1,0){\includegraphics[scale=1]{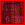}}
				\caption*{\fontfamily{Times New Roman}{FCTN}}
		\end{minipage}}
		\hspace{-1.7mm}
		\subfloat{
			\begin{minipage}[b]{0.121\linewidth}
				\captionsetup{
					labelfont=bf,        
					textfont={rm}, 
					singlelinecheck=true 
				}
				\includegraphics[width=1\linewidth]{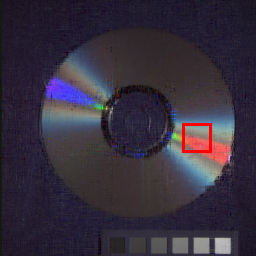}\vspace{1pt}
				\put(-25.1,0){\includegraphics[scale=1]{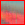}}\\
				\includegraphics[width=1\linewidth]{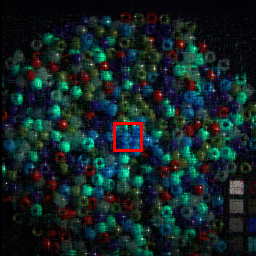}\vspace{1pt}
				\put(-25.1,0){\includegraphics[scale=1]{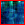}}\\
				\includegraphics[width=1\linewidth]{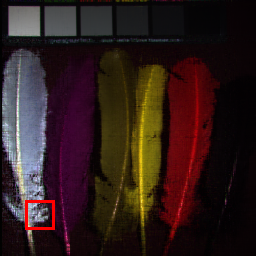}\vspace{1pt}
				\put(-25.1,0){\includegraphics[scale=1]{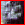}}\\
				\includegraphics[width=1\linewidth]{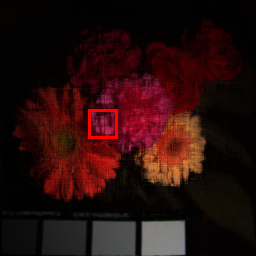}\vspace{1pt}
				\put(-25.1,0){\includegraphics[scale=1]{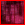}}
				\caption*{\fontfamily{Times New Roman}{HLRTF}}
		\end{minipage}}
		\hspace{-1.7mm}
		\subfloat{
			\begin{minipage}[b]{0.121\linewidth}
				\captionsetup{
					labelfont=bf,        
					textfont={rm}, 
					singlelinecheck=true 
				}
				\includegraphics[width=1\linewidth]{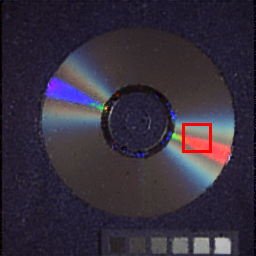}\vspace{1pt}
				\put(-25.1,0){\includegraphics[scale=1]{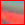}}\\
				\includegraphics[width=1\linewidth]{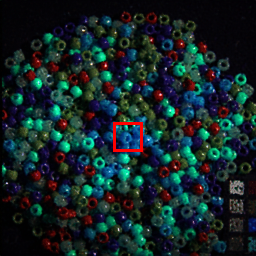}\vspace{1pt}
				\put(-25.1,0){\includegraphics[scale=1]{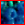}}\\
				\includegraphics[width=1\linewidth]{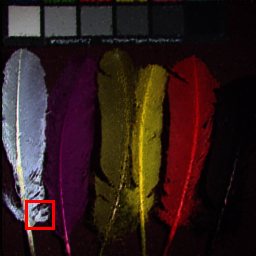}\vspace{1pt}
				\put(-25.1,0){\includegraphics[scale=1]{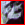}}\\
				\includegraphics[width=1\linewidth]{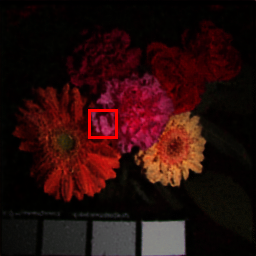}\vspace{1pt}
				\put(-25.1,0){\includegraphics[scale=1]{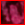}}
				\caption*{\fontfamily{Times New Roman}{DTR}}
		\end{minipage}}
		\hspace{-1.7mm}
		\subfloat{
			\begin{minipage}[b]{0.121\linewidth}
				\captionsetup{
					labelfont=bf,        
					textfont={rm}, 
					singlelinecheck=true 
				}
				\includegraphics[width=1\linewidth]{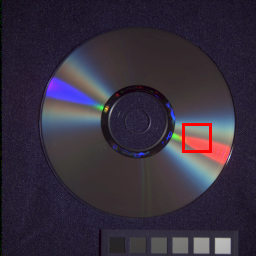}\vspace{1pt}
				\put(-25.1,0){\includegraphics[scale=1]{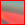}}\\
				\includegraphics[width=1\linewidth]{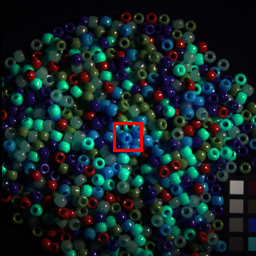}\vspace{1pt}
				\put(-25.1,0){\includegraphics[scale=1]{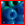}}\\
				\includegraphics[width=1\linewidth]{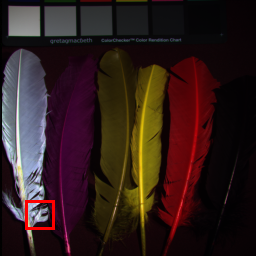}\vspace{1pt}
				\put(-25.1,0){\includegraphics[scale=1]{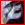}}\\
				\includegraphics[width=1\linewidth]{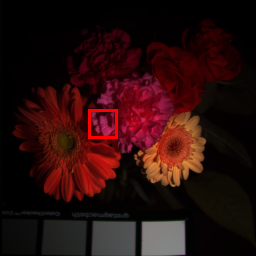}\vspace{1pt}
				\put(-25.1,0){\includegraphics[scale=1]{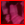}}
				\caption*{\fontfamily{Times New Roman}{Original}}
		\end{minipage}}
		\caption{The recovered pseudo-color images by different methods. From top to bottom: MSI\ \emph{Cd} (R:10, G:20, B:30), \emph{Beads} (R:10, G:20, B:30),  \emph{Feathers} (R:31, G:15, B:4), and \emph{Flowers} (R:31, G:15, B:4) with the tube missing (SR = 0.3).}	
		\label{visual4}
	\end{figure*}
	\begin{figure*}[!h]
		\centering
		\subfloat{
			\begin{minipage}[b]{0.121\linewidth}
				\captionsetup{
					labelfont=bf,        
					textfont={rm}, 
					singlelinecheck=true 
				}
				\includegraphics[width=1\linewidth]{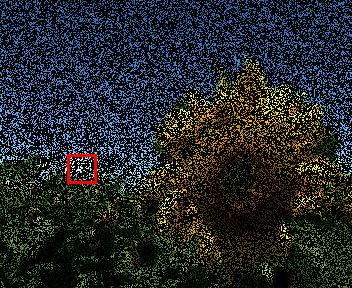}\vspace{1pt}
				\put(-25.1,0){\includegraphics[scale=1]{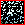}}\\
				\includegraphics[width=1\linewidth]{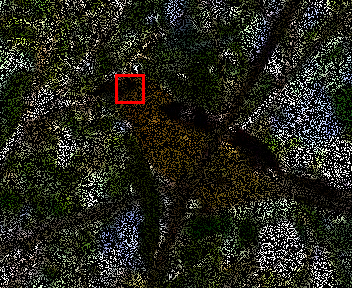}\vspace{1pt}
				\put(-25.1,0){\includegraphics[scale=1]{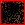}}
				
				\caption*{\fontfamily{Times New Roman}{Observed}}
		\end{minipage}}
		\hspace{-1.7mm}
		\subfloat{
			\begin{minipage}[b]{0.121\linewidth}
				\captionsetup{
					labelfont=bf,        
					textfont={rm}, 
					singlelinecheck=true 
				}
				\includegraphics[width=1\linewidth]{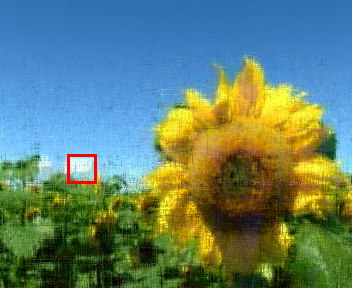}\vspace{1pt}
				\put(-25.1,0){\includegraphics[scale=1]{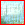}}\\
				\includegraphics[width=1\linewidth]{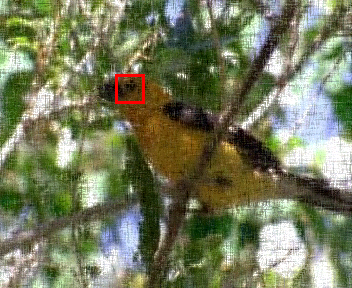}\vspace{1pt}
				\put(-25.1,0){\includegraphics[scale=1]{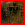}}
				
				\caption*{\fontfamily{Times New Roman}{TNN}}
		\end{minipage}}
		\hspace{-1.7mm}
		\subfloat{
			\begin{minipage}[b]{0.121\linewidth}
				\captionsetup{
					labelfont=bf,        
					textfont={rm}, 
					singlelinecheck=true 
				}
				\includegraphics[width=1\linewidth]{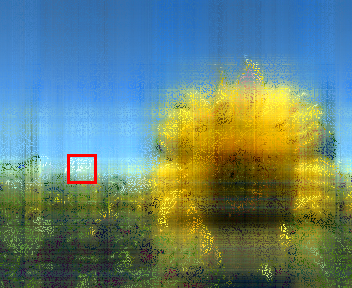}\vspace{1pt}
				\put(-25.1,0){\includegraphics[scale=1]{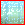}}\\
				\includegraphics[width=1\linewidth]{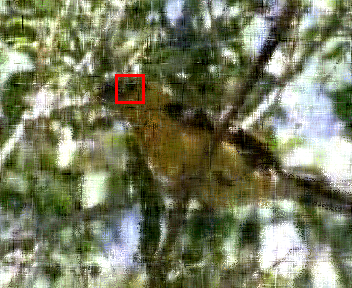}\vspace{1pt}
				\put(-25.1,0){\includegraphics[scale=1]{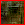}}

				\caption*{\fontfamily{Times New Roman}{TRLRF}}
		\end{minipage}}
		\hspace{-1.7mm}
		\subfloat{
			\begin{minipage}[b]{0.121\linewidth}
				\captionsetup{
					labelfont=bf,        
					textfont={rm}, 
					singlelinecheck=true 
				}
				\includegraphics[width=1\linewidth]{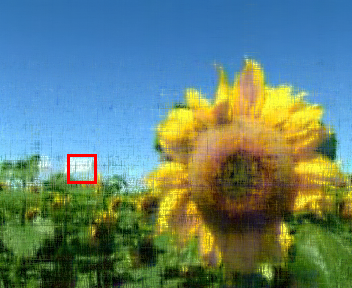}\vspace{1pt}
				\put(-25.1,0){\includegraphics[scale=1]{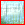}}\\
				\includegraphics[width=1\linewidth]{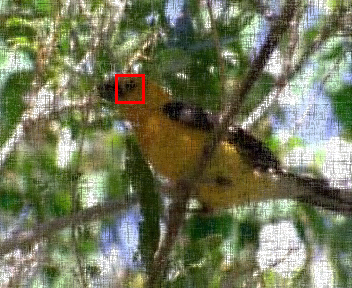}\vspace{1pt}
				\put(-25.1,0){\includegraphics[scale=1]{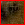}}
				\caption*{\fontfamily{Times New Roman}{FTNN}}
		\end{minipage}}
		\hspace{-1.7mm}
		\subfloat{
			\begin{minipage}[b]{0.121\linewidth}
				\captionsetup{
					labelfont=bf,        
					textfont={rm}, 
					singlelinecheck=true 
				}
				\includegraphics[width=1\linewidth]{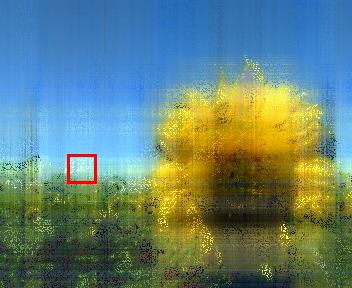}\vspace{1pt}
				\put(-25.1,0){\includegraphics[scale=1]{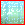}}\\
				\includegraphics[width=1\linewidth]{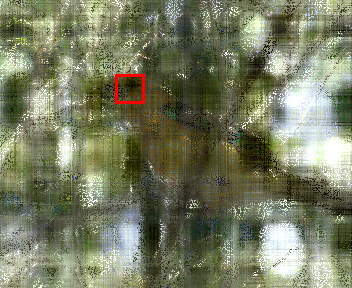}\vspace{1pt}
				\put(-25.1,0){\includegraphics[scale=1]{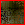}}
		
				\caption*{\fontfamily{Times New Roman}{FCTN}}
		\end{minipage}}
		\hspace{-1.7mm}
		\subfloat{
			\begin{minipage}[b]{0.121\linewidth}
				\captionsetup{
					labelfont=bf,        
					textfont={rm}, 
					singlelinecheck=true 
				}
				\includegraphics[width=1\linewidth]{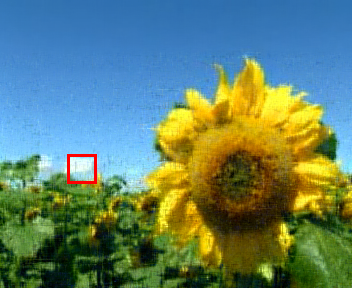}\vspace{1pt}
				\put(-25.1,0){\includegraphics[scale=1]{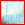}}\\
				\includegraphics[width=1\linewidth]{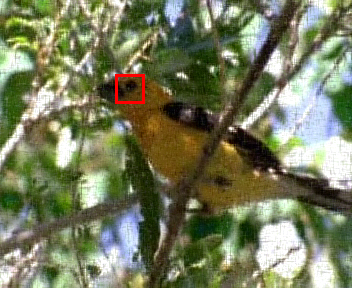}\vspace{1pt}
				\put(-25.1,0){\includegraphics[scale=1]{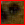}}
				
				\caption*{\fontfamily{Times New Roman}{HLRTF}}
		\end{minipage}}
		\hspace{-1.7mm}
		\subfloat{
			\begin{minipage}[b]{0.121\linewidth}
				\captionsetup{
					labelfont=bf,        
					textfont={rm}, 
					singlelinecheck=true 
				}
				\includegraphics[width=1\linewidth]{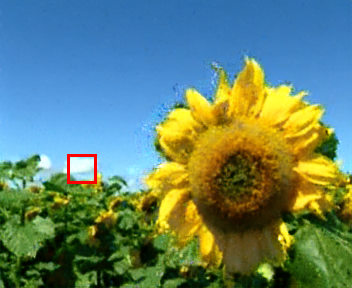}\vspace{1pt}
				\put(-25.1,0){\includegraphics[scale=1]{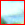}}\\
				\includegraphics[width=1\linewidth]{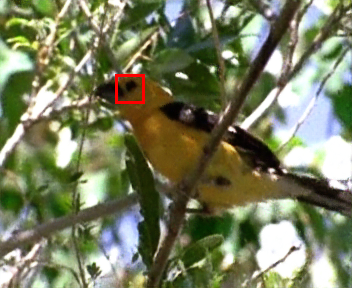}\vspace{1pt}
				\put(-25.1,0){\includegraphics[scale=1]{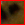}}
				\caption*{\fontfamily{Times New Roman}{DTR}}
		\end{minipage}}
		\hspace{-1.7mm}
		\subfloat{
			\begin{minipage}[b]{0.121\linewidth}
				\captionsetup{
					labelfont=bf,        
					textfont={rm}, 
					singlelinecheck=true 
				}
				\includegraphics[width=1\linewidth]{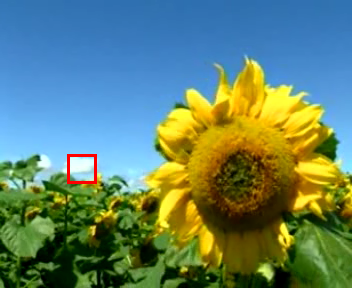}\vspace{1pt}
				\put(-25.1,0){\includegraphics[scale=1]{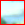}}\\
				\includegraphics[width=1\linewidth]{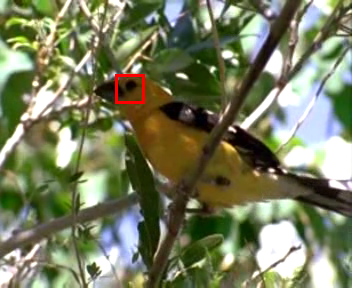}\vspace{1pt}
				\put(-25.1,0){\includegraphics[scale=1]{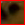}}
				
				\caption*{\fontfamily{Times New Roman}{Original}}
		\end{minipage}}
		\caption{The 10-th frame of the recovered results by all comparative methods on videos \emph{Sunflower} and \emph{Bird} with the tube missing (SR = 0.3).}	
		\label{visual5}
	\end{figure*}
	\subsection{Experimental Results}\label{Results}
	Tables~\ref{result1}-\ref{result2} list the evaluation metrics of the recovered results
	by different methods under the random missing and tube
	missing cases, respectively. Especially, we highlight best
	and second-best results by bold and underline, respectively. From
	Tables~\ref{result1}-\ref{result2}, we can observe that FTNN and FCTN perform well for random missing cases, while the TNN performs well for tube missing cases. The quantitative
	evaluation results of HLRTF are superior compared to TNN and FTNN for nearly all cases. The reason behind this is that the nonlinear transform in HLRTF obtained by learning a multi-layer neural network can better capture the frontal slice relationships of multi-dimensional
	data. Additionally, it is worth noting that DTR significantly outperforms HLRTF\footnote[6]{Due to the additional TV regularization term existing in HLRTF, some recovered results of HLRTF are better than DTR.} for most cases, which can be attributed to the powerful characterization ability of deep latent generator in contrast to shallow matrix factorization.
	
	To visually compare the recovered results, we present visual illustrations of the recovered results obtained by different methods in Figs. \ref{visual2}-\ref{visual5}.  Aiming
	at a better visual comparison, we mark one local area and enlarge it under
	each image. From Figs. \ref{visual2}-\ref{visual5}, we can observe that all the compared methods can somewhat recover the multi-dimensional data. The results of
	TNN and TRLRF contain some spatial blurring effects on MSIs (e.g., \emph{Cd} and \emph{Beads}).
	Although the FTNN and FCTN have better recovery effects than
	TNN and TRLRF visually, there are still some
	unsatisfactory performance in local details. The results of HLRTF
	have more spatial edges or textures compared
	to FTNN and FCTN, but small blurring effects still can
	be found. On the contrary, the proposed DTR	performs better on the preservation of global structure and local details compared with the comparative methods. 
	
	\raggedbottom
	Overall, the proposed DTR achieves superior performance
	on multi-dimensional data, including HSIs, MSIs, and videos. The superior performace of DTR can be
	attributed to the powerful representation ability of deep latent generative module and deep transform module. These two indispensable components
	allow DTR to more faithfully capture the rich textures and details in
	multi-dimensional data, which is particularly beneficial for multi-dimensional data
	recovery tasks.
	\section{Discussions}
 	Two indispensable components (i.e., the deep latent generative module and the deep transform module) in the proposed DTR framework complement to each other and work together to make up a unified framework. In this section, we discuss the roles of each component in the DTR framework.
	\label{section:6}
	\subsection{The Contribution of Deep Latent Generative Module $g_\theta(\cdot)$}
	\subsubsection{Deep Latent Generator vs  Shallow Matrix Factorization}
	To verify the characterization ability of deep latent generator, we compare the recovered results by the shallow characterization-based method (i.e., based on NMF) and deep characterization-based methods (i.e., based on deep latent generators). Table~\ref{result4} shows the quantitative comparison of the proposed DTR variants with different latent tensor characterizations for HSI \emph{Pavia}. 
	We can observe that the recovered results of shallow characterization-based method are generally inferior to that of deep characterization-based methods, and the number of network parameters is much larger. It is rational to say that the deep latent generator is more effective and efficient than shallow matrix factorization.
		\begin{table}[!h]\scriptsize
		\centering
		\caption{Quantitative comparison of the proposed DTR variants with different latent tensor characterizations for HSI \emph{Pavia} (random missing cases).}
		\setlength{\tabcolsep}{1.3mm}{
			\begin{tabular}{cccccccccc}
				\toprule
				\multicolumn{2}{c}{\multirow{3}{*}{Sampling rate}}&\multicolumn{2}{ c }{0.1}&\multicolumn{2}{ c }{0.2}&\multicolumn{2}{ c }{0.3}&\multirow{2}{*}{Params}\\
				\cmidrule{3-8}
				&&PSNR&SSIM&PSNR&SSIM&PSNR&SSIM&(M)\\
				\midrule
				Shallow &NMF&41.71&0.9845&46.75&0.9948&49.09&0.9968&2.102\\
				
				\midrule
				&V-Net&43.09&0.9880&44.79&0.9918&45.95&0.9933&0.165\\
				\multirow{2}{*}{Deep}&DenseNet&43.20&0.9881&48.80&0.9963&49.71&0.9971&0.659\\
				&U-Net&\underline{45.87}&\underline{0.9942}&\underline{48.85}&\underline{0.9966}&\underline{50.22}&\underline{0.9974}&0.565\\
				&U-Net-3D&\textbf{46.32}&\textbf{0.9945}&\textbf{49.33}&\textbf{0.9969}&\textbf{50.40}&\textbf{0.9975}&1.388\\
				\bottomrule
		\end{tabular}}\label{result4}
	\end{table}
	\subsubsection{The Influence of Different Networks}
Deep latent generative module plays a paramount role in the proposed DTR. 
We compare the recovered results by the proposed DTR with three classic networks, i.e., V-Net \cite{milletari2016v}, DenseNet \cite{huang2017densely}, and U-Net \cite{ronneberger2015u}. Table~\ref{result4} shows the quantitative comparison of different DTR variants for HSI \emph{Pavia}, where ``U-Net-3D" denotes the U-Net using 3D filters. 
We can observe that U-Net achieves better
results compared to V-Net and DenseNet, which
indicates that the U-Net may be a good choice for generating the latent tensor. In addition, while U-Net-3D achieves better
recovery results compared to U-Net, it comes with nearly
double number of parameters as U-Net. Considering
the trade-off between efficiency and performance, we set the $g_\theta(\cdot)$ as U-Net in our experiments.

	\subsubsection{The Influence of the Layer Number of $g_\theta(\cdot)$}
	The layer number of deep latent generative module $g_\theta(\cdot)$ is a crucial parameter that determines the characterization ability of $g_\theta(\cdot)$.  Table~\ref{result3}
	presents the quantitative comparison of the proposed DTR with
	different layer number $L$ of $g_\theta(\cdot)$ for HSI \emph{Pavia}. As shown in Table~\ref{result3}, we can observe that increasing $L$ enhances the
	performance when $L$ is small. However, when $L$ becomes
	larger, the results are not as desirable as expected. This is
	because a deeper network is more likely to suffer from the
	vanishing gradient. In our experiments, we set the layer number
	$L$ of $g_\theta(\cdot)$ to 2 to strike a balance
	between performance and the risk of vanishing gradient.
	\begin{table}[!h]\scriptsize
		\centering
		\caption{Quantitative comparison of the proposed DTR
			with different layer number (i.e., $L$ and $K$) of $g_\theta(\cdot)$ and $f_\xi(\cdot)$ for HSI \emph{Pavia} (random missing cases).}
		\setlength{\tabcolsep}{1.4mm}{
		\begin{tabular}{ccccccccc}
			\toprule
			\multicolumn{2}{c}{\multirow{3}{*}{Sampling rate}}&\multicolumn{2}{ c }{0.1}&\multicolumn{2}{ c }{0.2}&\multicolumn{2}{ c }{0.3}&\multirow{2}{*}{Params}\\
			\cmidrule{3-8}
			&&PSNR&SSIM&PSNR&SSIM&PSNR&SSIM&(M)\\
			\midrule
			&$L$ = 1&37.13&0.9555&39.27&0.9708&39.78&0.9743&0.095\\
			&$L$ = 2&45.87&\textbf{0.9942}&\textbf{49.32}&\textbf{0.9968}&\textbf{50.22}&\textbf{0.9974}&0.565\\
			$g_\theta(\cdot)$&$L$ = 3&\textbf{46.19}&\textbf{0.9942}&47.86&0.9958&49.58&0.9970&2.412\\
			&$L$ = 4&45.34&0.9935&48.28&0.9963&49.08&0.9968&9.821\\
			&$L$ = 5&45.28&0.9933&48.39&0.9962&49.04&0.9968&103.423\\
			\midrule
			&$K$ = 1&45.10&0.9929&48.97&0.9967&49.79&0.9973&0.559\\
			&$K$ = 2&\textbf{45.87}&\textbf{0.9942}&\textbf{49.32}&\textbf{0.9968}&\textbf{50.22}&\textbf{0.9974}&0.565\\
			$f_\xi(\cdot)$&$K$ = 3&45.49&0.9940&48.73&0.9965&49.81&0.9973&0.572\\
			&$K$ = 4&45.75&0.9939&48.66&0.9965&50.09&0.9973&0.578\\
			&$K$ = 5&45.46&0.9939&48.55&0.9965&49.54&0.9972&0.584\\
			\midrule
			\multicolumn{2}{ c }{DTR Without $f_\xi(\cdot)$}&44.36&0.9923&48.24&0.9961&49.74&0.9971&0.552\\
			\bottomrule
		\end{tabular}}\label{result3}
	\end{table}
\subsubsection{The Influence of $\hat{n}_3$ Value} 
To analyze the influence of $\hat{n}_3$ value on recovery effects of the proposed DTR, we conduct quantitative comparison of the proposed DTR with
different $\hat{n}_3$ value for HSI \emph{Pavia} in Table~\ref{result5}. As shown in Table~\ref{result5}, we can observe that a moderate $\hat{n}_3$ can obtain a satisfactory recovery effects. To balance the performance of the proposed DTR and the computational
complexity, we set $\hat{n}_3=n_3$ in all our experiments.
\begin{table}[!h]\scriptsize
	\centering
	\caption{Quantitative comparison of the proposed DTR
		with different $\hat{n}_3$ value for HSI \emph{Pavia} (random missing cases).}
	\setlength{\tabcolsep}{1.7mm}{
		\begin{tabular}{cccccccc}
			\toprule
			\multirow{3}{*}{Sampling rate}&\multicolumn{2}{ c }{0.1}&\multicolumn{2}{ c }{0.2}&\multicolumn{2}{ c }{0.3}&\multirow{2}{*}{Params}\\
			\cmidrule{2-7}
			&PSNR&SSIM&PSNR&SSIM&PSNR&SSIM&(M)\\
			\midrule
			$\hat{n}_{3}$ = $n_{3}/4$&44.95&0.9931&47.19&0.9956&49.02&0.9967&0.531\\
			$\hat{n}_{3}$ = $n_{3}/2$&44.99&0.9933&48.60&0.9964&50.27&0.9974&0.542\\
			$\hat{n}_{3}$ = $n_{3}$&\textbf{45.87}&\textbf{0.9942}&\textbf{49.32}&\textbf{0.9968}&50.22&0.9974&0.565\\
			$\hat{n}_{3}$ = $2n_{3}$&45.72&0.9941&48.97&\textbf{0.9968}&\textbf{50.40}&\textbf{0.9975}&0.611\\
			\bottomrule
	\end{tabular}}\label{result5}
\end{table}
	\subsection{The Contribution of Deep Transform Module $f_\xi(\cdot)$}
The deep transform module $f_\xi(\cdot)$ is another indispensable component in the
proposed DTR.
Table~\ref{result3} presents the quantitative comparison of the proposed
DTR with different the layer number $K$ of $f_\xi(\cdot)$ for HSI
\emph{Pavia}. It can be observed that using the deep transform module can achieve better recovered results, which reflects the effectiveness of deep transform module. Particularly, when $K$ = 1, $f_\xi(\cdot)$ degenerates into a linear transform. We can observe that nonlinear transform (i.e., $K\geq2$) can achieve better recovered results, which
indicates that nonlinear transform can better capture the frontal slice relationships of multi-dimensional data. Similar to the deep latent generative module, we can observe that increasing $K$ enhances the performance when $K$ is small.
However, when $K$ becomes larger, the performance tends
to drop. Thus, in our
experiments, we set the layer number $K$ of $f_\xi(\cdot)$ to 2.
%
	\section{Conclusion}
	\label{section:7}
	In this paper, we proposed a unified deep tensor representation	(termed as DTR) framework by synergistically combining the deep latent generative module and the deep transform module. The new DTR framework not only allows us to better understand the classic shallow representations, but also leads us to explore new representation. Extensive experiments demonstrate that the
	proposed DTR can achieve satisfactory performance on multi-dimensional
	data recovery problems.




%

\bibliographystyle{IEEEtran}
\bibliography{egbib}

\vfill

\end{document}